\documentclass[sigconf,nonacm]{acmart}

\usepackage{enumitem}
\usepackage{subcaption}
\usepackage{newfloat}
\usepackage{listings}
\usepackage{cleveref}
\usepackage{titletoc}

\DeclareCaptionStyle{ruled}{labelfont=normalfont,labelsep=colon,strut=off}
\floatstyle{ruled}
\newfloat{listing}{tb}{lst}{}
\floatname{listing}{Listing}

\definecolor{ResultColor}{RGB}{0,90,156}
\newcounter{resultctr}
\newcommand{\Result}[1]{%
  \refstepcounter{resultctr}%
  \leavevmode\textcolor{ResultColor}{\textbf{Result~\theresultctr:}}\ \textit{#1}\ }

\AtBeginDocument{%
  }

\setcopyright{acmlicensed}
\copyrightyear{2025}
\acmYear{2025}
\acmDOI{XXXXXXX.XXXXXXX}
\acmConference[KDD '25]{Proceedings of the 31st ACM SIGKDD Conference on
  Knowledge Discovery and Data Mining}{August 03--07, 2025}{Toronto, Canada}
\acmISBN{978-1-4503-XXXX-X/2025/08}

\begin{document}

\title{ASAP: \underline{A}gent-\underline{S}ystem Co-Design for Wall-Clock-Centered \underline{A}uto H\underline{P}O Research for ML Experiments}

\author{Taicheng Guo}
\authornote{Key Contributors.}
\affiliation{\institution{University of Notre Dame}\country{United States}}
\author{Haomin Zhuang}
\authornotemark[1]
\affiliation{\institution{University of Notre Dame}\country{United States}}
\author{Kehan Guo}
\affiliation{\institution{University of Notre Dame}\country{United States}}
\author{Yujun Zhou}
\affiliation{\institution{University of Notre Dame}\country{United States}}
\author{Nitesh V. Chawla}
\affiliation{\institution{University of Notre Dame}\country{United States}}
\author{Olaf Wiest}
\affiliation{\institution{University of Notre Dame}\country{United States}}
\author{Xiangliang Zhang}
\authornote{Corresponding author.}
\affiliation{\institution{University of Notre Dame}\country{United States}}

\renewcommand{\shortauthors}{Guo et al.}

\begin{abstract}
Hyperparameter Optimization (HPO) is essential for maximizing machine learning (ML) model performance, and its core challenge is sample efficiency: finding strong configurations within a limited budget. Because every HPO tool relies on a surrogate prior that imparts its own inductive bias, individual tools tend to struggle once problems become sufficiently diverse and drift from these priors. Motivated by the strong reasoning and generalization capabilities of LLMs, recent work has explored using LLMs for HPO and reports improved per-iteration performance. Yet these methods share two limitations with a common origin: they use the LLM as a \emph{single-tool replacement evaluated by iteration count}. (i) The LLM is designed to be deployed in place of, rather than alongside, prior tools, and is itself constrained by its pretraining objective to a particular family of inductive-biased proposals; the resulting single-source setup still does not handle the full diversity of problems. (ii) Per-iteration evaluation silently ignores that, in real runs, LLM inference or tool execution is paid serially on top of the model evaluation every round, so iteration-count gains do not translate into real \textbf{end-to-end wall-clock} gains.
We present \textbf{ASAP}, an agent--system co-design that addresses both limitations. On the \emph{agent} side, ASAP uses the LLM to integrate a diverse pool of inductive-biased optimizers and to select among their proposals each round. On the \emph{agent--system co-design} side, ASAP re-architects the loop to reduce end-to-end wall-clock while preserving regret quality: a \emph{prefix-stable prompt} maximizes KV-cache reuse across rounds; \emph{speculation parallelism} hides the remaining LLM and tool latency under model evaluation via a relative-error accept test; and a \emph{Self-Tuner} adapts the speculation threshold from execution logs off the critical path. Extensive experiments on diverse modern HPO tasks
demonstrate that ASAP consistently outperforms baselines, underscoring the value of tool integration and agent-system co-design.
\end{abstract}


\maketitle

\section{Introduction}
\label{Sec:Introduction}

Hyperparameter Optimization (HPO) is essential for maximizing the performance of machine learning (ML) models, since the choice of hyperparameters strongly affects model effectiveness. HPO is inherently a sequential decision-making process: at each iteration, given past observations (hyperparameters and corresponding performance), the optimization algorithm selects the next set of hyperparameters for evaluation~\cite{JMLR:v13:bergstra12a}. Existing HPO methods build a surrogate model to approximate the black-box objective of the target ML model, a candidate sampler to propose configurations, and an acquisition function to choose among them. They fall into two primary categories: (1) \emph{Conventional statistical} approaches such as Scikit-Optimize (SKopt)~\cite{skopt} and Optuna~\cite{akiba2019optunanextgenerationhyperparameteroptimization}, which use Gaussian processes~\cite{NIPS1995_7cce53cf} or Tree-structured Parzen Estimators (TPE)~\cite{watanabe2026treestructuredparzenestimatorunderstanding} as surrogates, random or Bayesian sampling~\cite{NIPS2014_b610047c} as candidates, and expected improvement or upper confidence bound for selection; and (2) \emph{LLM-based} approaches~\cite{liu2024largelanguagemodelsenhance, liu2025largelanguagemodelagent}, in which an LLM directly plays one or more of these roles given past observations.

\begin{figure}[t]
    \centering
\includegraphics[width=\columnwidth]{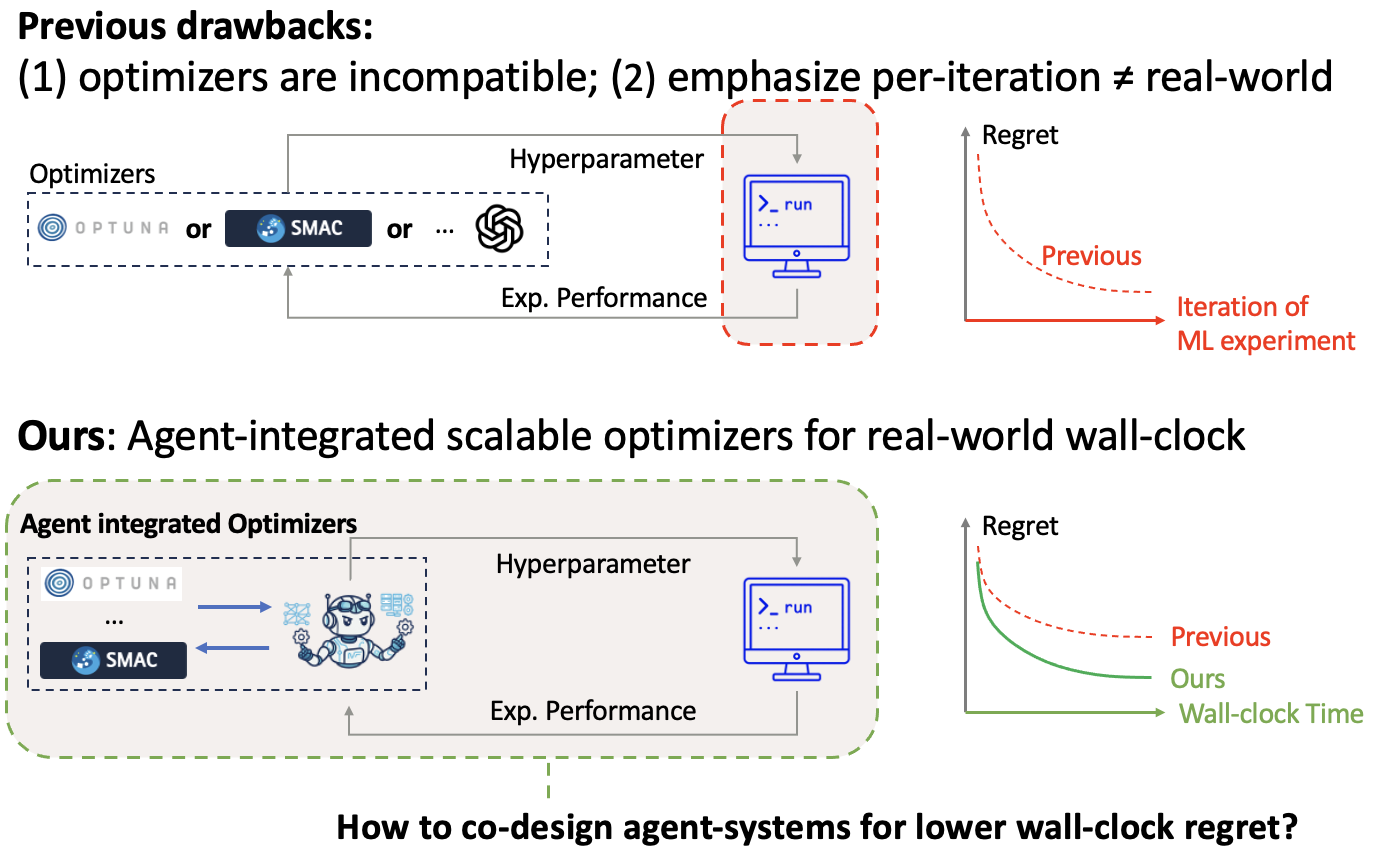}
    \caption{Motivation of ASAP. Prior LLM-HPO replaces a \emph{single} optimizer and is judged on per-iteration regret, ignoring the overhead paid; ASAP instead integrates a scalable pool of optimizers under one LLM agent and co-designs the system for low regret per unit of real-world wall-clock.}
    \label{fig:intro}
\end{figure}

Real-world HPO tasks vary substantially in data distribution and model architecture, and \textbf{every HPO tool relies on a surrogate prior that imparts its own inductive bias}. For instance, Gaussian-process Bayesian optimization assumes a smooth, GP-distributed response, and TPE factors the search through a good/bad density split over past observations. Each such prior yields strong proposals when it aligns with the underlying optimization landscape, but the same alignment also causes the tool to struggle once the next task drifts away from those assumptions~\cite{li2024studybayesianneuralnetwork}. With the stronger reasoning and generalization capabilities of LLMs, LLM-based optimizers such as LLAMBO~\cite{liu2024largelanguagemodelsenhance} have recently been proposed and inherit priors implicit in the LLM's pretraining corpus. However, they do not escape this trap: by design they replace a conventional tool with the LLM, and the LLM itself is constrained by its pretraining objective to a particular family of inductive-biased proposals, so a single LLM-as-tool setup is still bound to one source of inductive bias. Another ignored drawback is that most methods evaluate HPO almost exclusively against iteration count, while in real runs LLM inference or tool execution is paid serially on top of the model evaluation $T_{\mathrm{ML}}$ every round; summed across iterations this overhead can rival or exceed the cost the optimizer is trying to amortize, so iteration-count gains do not translate into \textbf{end-to-end wall-clock} gains. Overall, we summarize the two drawbacks as: (1) LLM and prior tools are each bound to their own inductive bias and are designed to substitute rather than compose, leaving the system fragile when problems drift from a single prior; and (2) prior systems are iteration-count-centric, ignoring the LLM and tool overhead that dominates real wall-clock.

To address these drawbacks, our key insights and design principles are twofold. (1) \textbf{Integration over replacement}: rather than betting on any one inductive-biased proposer, we ask whether a diverse pool of such proposers, statistical tools and LLM-as-tool alike, can be made compatible under one LLM coordinator that scales to any new tool. While integration does not eliminate inductive bias, combining complementary biases is a natural lever for mitigating per-tool fragility and yielding stable, robust optimization across diverse problems. (2) \textbf{Wall-clock as the design objective}: the honest axis for HPO is end-to-end wall-clock, not iteration count, and once wall-clock is the target the prompt structure, tool scheduling, and execution policy become coupled and can no longer be tuned in isolation; they must be designed together with the agent itself.

Guided by these principles, we present \textbf{ASAP}, an agent-system co-design built on an integrated pool of HPO optimizers and organized into two modules. (1) \textbf{Agent-integrated optimizers}: each round, a diverse pool of optimizer tools proposes hyperparameter candidates in parallel, and an LLM-as-Judge selects among them, so the pool's complementary inductive biases jointly cover problems that no single proposer covers alone. (2) \textbf{Agent-system co-design for wall-clock}: we profile the full optimization pipeline and identify two dominant cost sources, hyperparameter proposal and ML-experiment execution, and design against each. A \emph{KV-Cache-Aware prefix-stable prompt} fixes an immutable task/tool/rule prefix so the LLM's per-round prefill cost collapses toward suffix-only via KV-cache reuse. A \emph{speculation parallelism} mechanism predicts the current round's ML evaluation outcome with the LLM, speculatively runs the next round's tools and Judge during the current evaluation, and commits the speculative result under a relative-error accept test that never re-runs the dominant model evaluation. A \emph{Self-Tuner} runs off the critical path as a background process, adapting the speculation threshold from execution logs so the system improves itself during optimization. Together, these mechanisms preserve the selection quality of LLM coordination while substantially reducing regret per unit wall-clock.

The contributions of this work are summarized as follows:
\begin{itemize}[nolistsep, leftmargin=*]
    \item New problem framing: to our knowledge, this is the first systematic study of wall-clock-centered LLM agent-system co-design for HPO. We identify two ignored drawbacks of prior methods: (1) LLM and tools are each inductive-biased and were designed to substitute rather than compose, so the system struggles once problems drift; (2) prior systems are iteration-count-centric, ignoring the LLM and tool overhead that dominates real wall-clock.
    \item ASAP framework: we propose ASAP, an agent-system co-design with two coupled modules: (1) agent-integrated optimizers, where complementary biases jointly cover diverse problems; and (2) agent-system co-design for wall-clock, which jointly designs the prompt, speculation parallelism, and a Self-Tuner to reduce wall-clock time while preserving regret.
    \item Empirical validation: extensive quantitative and qualitative experiments on tasks spanning XGBoost, RandomForest, and vision (Wide-ResNet/ResNet/CNN), language (Transformer), and protein (Transformer) models across different datasets demonstrate the superior effectiveness and efficiency of ASAP. An in-depth analysis -- in particular a landscape analysis -- further shows that ASAP is uniformly robust across rugged, multi-modal, deceptive, and anisotropic landscape axes, the regime where single-bias BO tools each drop on at least one axis.
\end{itemize}

\section{Related Work}
\label{Sec:related_work}

\subsection{LLMs for AutoML and HPO}
With the rapid advancement of LLMs, there is a growing research trend in leveraging them for autonomous machine learning (AutoML)~\cite{tornede2024automlagelargelanguage}. However, most LLM-based AutoML approaches focus on broader tasks, such as dataset processing and workflow orchestration~\cite{hong2024datainterpreterllmagent, guo2024dsagentautomateddatascience, guo2025mtsql, chi2024selatreesearchenhancedllm}, while overlooking HPO. Typically, they allow LLMs to directly generate hyperparameter settings without optimization, which can prevent ML models from reaching their best performance.
Among LLM-HPO methods, LLAMBO~\cite{liu2024largelanguagemodelsenhance} uses an LLM as both the surrogate model and the acquisition function for HPO, while other works~\cite{zhang2024usinglargelanguagemodels, liu2025largelanguagemodelagent, guo2026autollmresearch} introduce LLMs as agents that analyze experimental results and directly suggest hyperparameters for subsequent iterations. In all of these works, the LLM is \emph{only designed to replace a single} HPO tool, and ignores the LLM inference cost paid serially every round.
\textit{In contrast, ASAP (i) integrates a \emph{pool} of HPO tools under one LLM-as-Judge to mitigate the inductive bias of any single tool, and (ii) is designed for \emph{end-to-end wall-clock time}, hiding the agent cost rather than paying it serially.}

\subsection{LLM Agent-System Co-Design}
\paragraph{Speculative Acceleration for LLM Agents}
Speculation was originally introduced for decoding acceleration, where a draft
model proposes tokens verified in parallel under a distribution-preserving rule
\citep{leviathan2023fastinferencetransformersspeculative}. As tasks become long-horizon and LLM agents
are widely deployed, recent work lifts speculation to the agent loop: it
speculates next actions or tool calls executed ahead of verification
\citep{ye2026speculativeactionslosslessframework,guan2025dynamicspeculativeagentplanning}, and downstream workflow nodes verified
asynchronously with rollback \citep{ro2025sherlockreliableefficientagentic}. These methods all speculate \emph{discrete} targets under exact, distributional, or binary acceptance tests.
\textit{ASAP differs in object and test: it speculates a \emph{continuous}
model-performance value that drives tool inputs, and commits under a
relative-error gate whose fallback re-runs only
the cheap tool+Judge stage on rejection. The worst-case round therefore equals the non-speculative baseline, while accepted rounds hide the LLM+tool overhead under the model evaluation.}

\paragraph{Self-Tuning/Self-Evolving Agents}
A growing body of work studies agents that adapt their own behavior from experience~\cite{wu2026evolverselfevolvingllmagents, chen2025multiagentevolvellmselfimprove, gao2026surveyselfevolvingagentswhat}. Building on this self-evolving paradigm, ASAP's Self-Tuner runs \emph{as a background process}, so its latency is never paid on the critical path. By continuously analyzing runtime logs, it adapts the speculation acceptance threshold $\tau$ -- the single knob that trades regret against the wall-clock saved by speculative acceleration -- and adjusts it dynamically for better runtime decisions.

\section{ASAP Module 1: Agent-Integrated Tools}
\label{sec:c1-judge}

\begin{figure*}[t]
    \centering
    \includegraphics[width=0.8\textwidth]{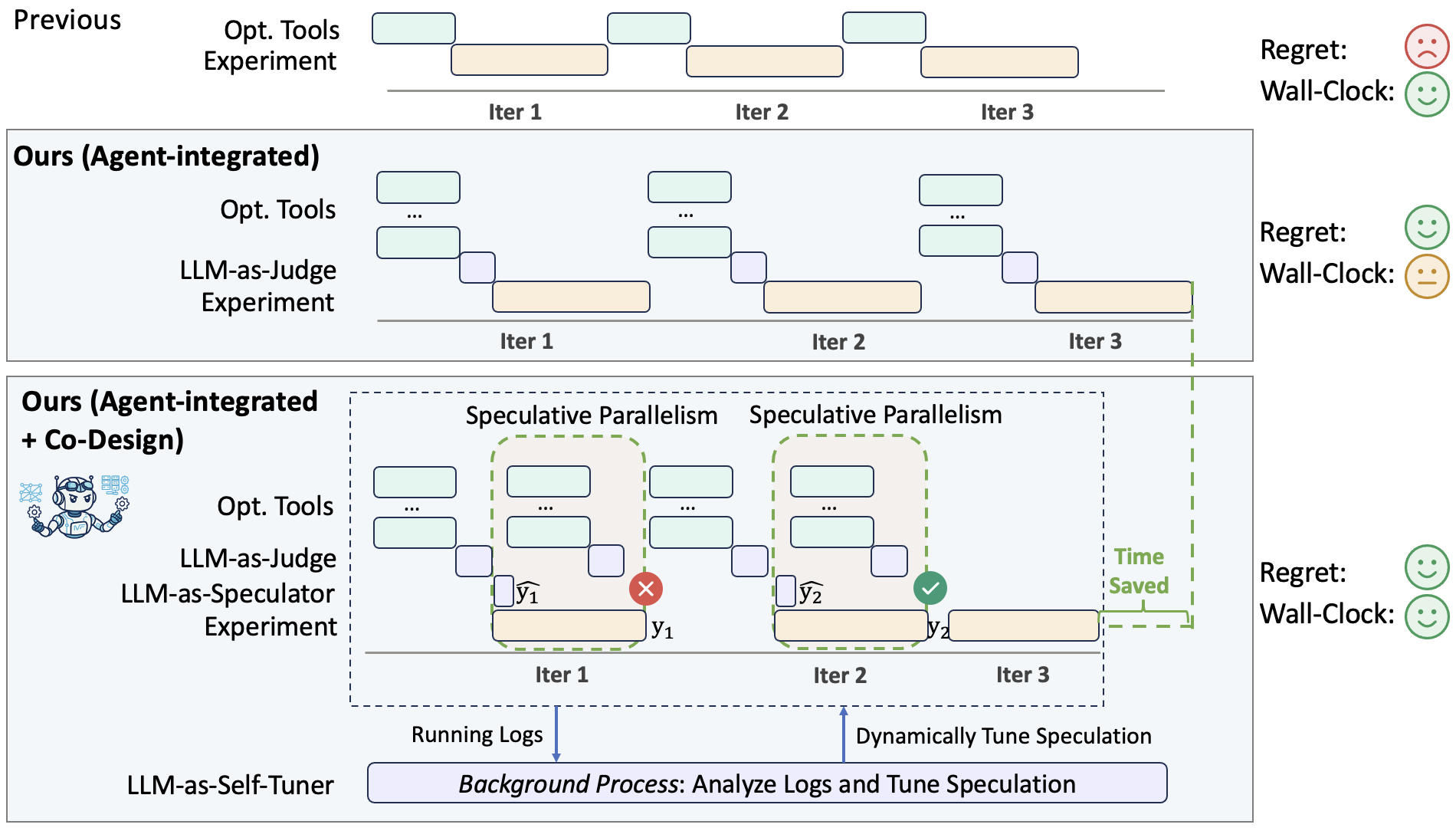}
    \caption{ASAP execution loop. After a serial warm-up, the Speculator predicts round $n$'s outcome $\hat{y}_n$, and the next round's tools and Judge run speculatively on $\hat{y}_n$ while round $n$'s ML evaluation ($T_{\text{ML}}$) is in flight. On accept ($|\hat{y}_n - y_n|/|y_n| < \tau$), the speculative work commits, hiding the LLM+tool cost under $T_{\text{ML}}$; on reject, the tool+Judge stage re-runs with the true $y_n$, matching the non-speculative wall-clock for that round. The Self-Tuner updates $\tau$ in the background.}
    \label{fig:asap_architecture}
\end{figure*}

\subsection{Why Integrate Rather Than Replace}
\label{sec:integrate-vs-replace}

\paragraph{Setup and notation.}
Let $f:\mathcal{X}\to\mathbb{R}$ denote the black-box objective of an ML task and $D_n=\{(x_t,y_t)\}_{t=0}^{n-1}$ the observation history available at the start of round $n$, with $y_t=f(x_t)$; round $n$ then proposes the $n$-th point $x_n$. Each HPO tool $i$ is characterized by an inductive bias $\theta_i$ that parameterizes a surrogate prior $p_i(f;\,\theta_i)$ over the objective. Conditioning on the observations yields the posterior $p_i(f\mid D_n;\,\theta_i)$ and an acquisition policy $\pi_i(\cdot\mid D_n;\,\theta_i)$ over $\mathcal{X}$; the round-$n$ proposal $g_i(D_n;\theta_i)\in\mathcal{X}$ is the acquisition maximizer of $\pi_i$ (or a draw from $\pi_i$ for stochastic proposers such as TPE). Tools differ precisely in this prior, encoded by $\theta_i$: GP-based Bayesian optimization assumes a smooth, stationary kernel~\cite{snoek2012practicalbayesianoptimizationmachine}; TPE factors the search through a good/bad density split over past observations~\cite{NIPS2011_86e8f7ab}; SMAC uses a random-forest surrogate over structured spaces~\cite{lindauer2022smac3versatilebayesianoptimization}; and LLM-as-proposer methods such as LLAMBO~\cite{liu2024largelanguagemodelsenhance} inherit a prior implicit in the LLM's pretraining corpus.

\paragraph{Single-tool fragility and the grand-mean floor.}
In practice, an HPO service must optimize not a single task but a \emph{suite} of diverse tasks $\mathcal{F}=\{f^{(1)},\dots,f^{(M)}\}$ spanning different model architectures, search spaces, and data modalities. Let the simple regret of running tool $i$ alone for $n$ rounds on task $f$ be
\begin{equation}
  R_i(f,n) \;=\; f^{*} \,-\, \max_{t\le n} f(x_t),
  \quad x_t \sim \pi_i(\cdot\mid D_{t};\theta_i),
  \label{eq:simple-regret}
\end{equation}
where $f^{*}=\max_{x\in\mathcal{X}} f(x)$ is the global optimum of $f$, $\pi_i(\cdot\mid D_{t};\theta_i)$ is tool $i$'s acquisition policy under prior $\theta_i$, and $D_{t}$ is the history available at round $t$. A prior $\theta_i$ that matches the structure of $f$ concentrates $\pi_i$ on promising regions and yields small $R_i$, but the same concentration becomes pathological when the structure of $f$ departs from what $\theta_i$ assumes: a GP with a stationary kernel may excel on smooth objectives yet squander most of its budget on rugged combinatorial landscapes where TPE or SMAC would thrive, and vice versa. Because $\mathcal{F}$ spans heterogeneous task structures, any fixed prior $\theta_i$ is necessarily mismatched on part of $\mathcal{F}$: whenever two tasks are best served by incompatible priors, no single $\theta_i$ can be optimal on both. Hence no fixed $\theta_i$ minimizes $R_i(f,n)$ uniformly over $\mathcal{F}$.\footnote{This echoes the classical No-Free-Lunch results for optimization~\cite{wolpert2002no}, although those require the function class to be closed under permutation, a condition that the structured objectives considered here do not satisfy.}

Aggregating over the suite, define $\bar R_i(\mathcal{F},n)=\tfrac{1}{M}\sum_{f\in\mathcal{F}} R_i(f,n)$ as tool $i$'s mean regret on $\mathcal{F}$. A practitioner who does not know in advance which $\theta_i$ matches which task must fall back on a task-agnostic choice; the canonical baseline is to pick a tool uniformly at random, whose expected mean regret is the \emph{grand-mean regret}
\begin{equation}
  \bar R(\mathcal{F},n) \;\triangleq\; \frac{1}{K}\sum_{i=1}^{K} \bar R_i(\mathcal{F},n),
  \label{eq:grand-mean}
\end{equation}
where $K$ is the number of available tools and $M=|\mathcal{F}|$ is the number of tasks. The limitation here is epistemic rather than one of raw capability: a tool whose prior matches $f$ can of course beat $\bar R(\mathcal{F},n)$ on that task, but without knowing in advance which $\theta_i$ matches which $f$, every task-agnostic selection rule has expected mean regret exactly $\bar R(\mathcal{F},n)$. This uninformed-selection baseline, not any single tool's favorable case,
is what any task-agnostic strategy is structurally stuck at.
The analysis below shows that \emph{integration plus an informed
Judge} escapes the per-round analogue of this baseline (Eq.~\eqref{eq:condition}),
and Sec.~\ref{Sec:exp} shows this per-round escape carries through to lower
trajectory-level regret in practice.

LLM-as-proposer methods do not escape it. Their design target is to replace $\pi_i$ with an LLM-based proposal $\pi_{\mathrm{LLM}}(x\mid D_n)$, but the LLM is itself a fixed-bias proposer: next-token pretraining collapses $\pi_{\mathrm{LLM}}$ into its own implicit prior $\theta_{\mathrm{LLM}}$, so it disproportionately suggests learning rates near $10^{-3}$ or batch sizes that are powers of two even when the task favors otherwise. Substituting one tool for another therefore moves along the $\theta$ axis but does not leave it: $\bar R_{\mathrm{LLM}}(\mathcal{F},n)$ is just another single-tool mean regret, with no structural reason to fall below the uninformed-selection baseline $\bar R(\mathcal{F},n)$ across diverse $\mathcal{F}$.

\paragraph{Integration with a random Judge recovers the grand-mean baseline.}
We now show that integration is the right escape, provided one mild condition holds. At round $n$, given the shared history $D_n$, each tool $i\in[K]$ proposes a batch $G_i(D_n;\theta_i)\subset\mathcal{X}$, and the candidate set is their union
\begin{equation}
  C_n \;=\; \bigcup_{i\in[K]} G_i(D_n;\theta_i),
  \label{eq:candidate-set}
\end{equation}
where $[K]=\{1,\dots,K\}$ indexes the active pool. We treat $C_n$ as an indexed family so that coincident proposals are counted with multiplicity, and write $|C_n|$ for its size (here $|C_n|\approx 18$, since one tool emits a batch of candidates and the rest one each). A Judge $h$ with parameters $\phi$ then selects $x_n=h(C_n,D_n;\phi)$. As a thought-experiment baseline, consider the \emph{naive Judge} that draws a candidate $c\sim\mathrm{Unif}(C_n)$.

\noindent\textbf{Proposition (Random-assignment equivalence).}
Let $r_c(f,n)=f^{*}-f(c)$ denote the per-round regret of candidate $c\in C_n$ on task $f$ under the shared history $D_n$, and let
\begin{equation}
  \bar r(\mathcal{F},n) \;\triangleq\; \frac{1}{|C_n|\,M}\sum_{c\in C_n}\sum_{f\in\mathcal{F}} r_c(f,n)
  \label{eq:perround-grandmean}
\end{equation}
be the \emph{per-round grand mean}, the per-round, shared-history counterpart of $\bar R(\mathcal{F},n)$ in Eq.~\eqref{eq:grand-mean}. If the naive Judge picks uniformly at random from $C_n$, its expected per-round mean regret over $\mathcal{F}$ equals this per-round grand mean:
\begin{align}
  \frac{1}{M}\sum_{f\in\mathcal{F}}\!
    \mathbb{E}_{x_n\sim\mathrm{U}(C_n)}\!\bigl[f^{*}-f(x_n)\bigr]
  &\;=\; \frac{1}{|C_n|\,M}\sum_{c\in C_n}\sum_{f\in\mathcal{F}} r_c(f,n) \notag \\
  &\;=\; \bar r(\mathcal{F},n),
  \label{eq:random-equiv}
\end{align}
where $\mathrm{U}(C_n)$ is the uniform draw over the $|C_n|$ candidates in $C_n$ (counted with multiplicity). In words, integrating tools and then selecting uniformly at random recovers exactly the pool-average regret that uninformed selection already incurs on diverse $\mathcal{F}$, so \textbf{integration alone is not enough}; the gain must come from the Judge.

\paragraph{From random to informed: the condition for integration to win.}
Equation~\eqref{eq:random-equiv} also reveals the path forward. Let
\begin{equation}
  \bar r_{\mathrm{int}}(\mathcal{F},n)
  \;=\; \frac{1}{M}\sum_{f\in\mathcal{F}}\!\bigl[f^{*}-f(x_n)\bigr],
  \;\; x_n \!=\! h(C_n,D_n;\phi),
  \label{eq:judge-regret}
\end{equation}
denote the integrated agent's per-round mean regret with a real (non-random) Judge $h$. Combining Eqs.~\eqref{eq:random-equiv} and~\eqref{eq:judge-regret}, both per-round, shared-history quantities, gives the criterion
\begin{equation}
  \bar r_{\mathrm{int}}(\mathcal{F},n) \;<\; \bar r(\mathcal{F},n)
  \;\;\Longleftrightarrow\;\;
  h \text{ outranks } \mathrm{U}(C_n),
  \label{eq:condition}
\end{equation}
where the right-hand side reads ``$h$ selects from $C_n$ better than uniform random.'' \textbf{What the criterion does and does not claim.}
We stress that Eq.~\eqref{eq:condition} is a \emph{per-round, shared-history}
statement: both $\bar{r}_{\mathrm{int}}(\mathcal{F},n)$ and $\bar{r}(\mathcal{F},n)$
are computed on the \emph{same} committed history $D_n$ that the integrated agent
has already built, and compare the Judge against uniform selection over the
\emph{same} candidate set $C_n$. Under this criterion the Judge need not be optimal:
clearing this single, deliberately mild bar suffices for the integrated agent to
beat uniform selection over $C_n$ at round $n$. This is weaker than, and should not
be conflated with, a claim about the trajectory-level grand mean
$\bar{R}(\mathcal{F},n)$ in Eq.~\eqref{eq:grand-mean}, where each tool runs an entire
trajectory on its own history; we do not prove in closed form that a per-round win
composes into a trajectory-level win, since the histories themselves diverge once
selection begins. What we \emph{do} claim, and verify empirically
(Sec.~\ref{sec:rq1-eff}, Fig.~\ref{fig:judge-vs-random}), is that the per-round
precondition Eq.~\eqref{eq:condition} holds at \emph{every} round, and that the
resulting committed trajectory attains lower end-to-end regret than every baseline
(RQ1). The per-round criterion is thus the \emph{mechanism} we can characterize
cleanly; its compounding into the global regret gain is an empirical result, not a
theorem.

Two facts make this criterion load-bearing rather than tautological. First,
Eq.~\eqref{eq:random-equiv} shows that integration \emph{alone}---pooling tools and
selecting at random---merely recovers the pool-average regret that uninformed
selection already incurs, so the entire gain must come from the Judge, not from
having more tools. Second, the Judge's task is \emph{discriminative}, not
generative: it only ranks the small, historically grounded set $C_n$, whereas a
proposer must generate good points over all of $\mathcal{X}$. ASAP thereby converts
a hard generative inductive bias into a far milder discriminative one, which is why
the criterion in Eq.~\eqref{eq:condition} is routinely satisfiable. Moreover, the
more heterogeneous $\mathcal{F}$ becomes, the larger the gap between the per-round
grand mean and the per-task best, so the problem grows harder for any task-agnostic
strategy while the Judge's task (ranking the same $|C_n|$ candidates) does not.

\subsection{How Integrating: The Integrated Loop}
\label{sec:integrated-loop}

\paragraph{Tool pool and per-round procedure.}
ASAP instantiates the candidate family $C_n$ from a deliberately diverse pool of $K=8$ optimizers: seven statistical tools, namely GP~\cite{snoek2012practicalbayesianoptimizationmachine}, TPE~\cite{NIPS2011_86e8f7ab}, SKOPT~\cite{skopt}, TuRBO~\cite{eriksson2020scalableglobaloptimizationlocal}, Optuna~\cite{akiba2019optunanextgenerationhyperparameteroptimization}, SMAC~\cite{lindauer2022smac3versatilebayesianoptimization}, and DNGO~\cite{snoek2015scalablebayesianoptimizationusing}, together with one LLM-based optimizer, LLAMBO~\cite{liu2024largelanguagemodelsenhance}. Each tool $i$ proposes a small batch of candidates $G_i(D_n;\theta_i)\subset\mathcal{X}$, and the round-$n$ candidate set is their union $C_n \;=\; \bigcup_{i\in[K]} G_i(D_n;\theta_i)$. The pool is intentionally extensible: a more diverse pool widens the gap between uniform selection and the per-task best, i.e.\ the headroom of Eq.~\eqref{eq:condition} that a better-than-random Judge converts into gains.
Given an ML task $f$ and its textual problem context (an example Judge prompt is shown in Table~\ref{tab:judge-prompt}), each round $n$ proceeds as follows. \textbf{(1)} Every tool $i\in[K]$ proposes a candidate $g_i(D_n;\theta_i)\in\mathcal{X}$ \textbf{in parallel}; the tools share only the history $D_n$ and have no inter-tool dependency, so the per-round propose \emph{latency} is that of the slowest tool rather than the sum over the $K$ tools. \textbf{(2)} The LLM-as-Judge selects $x_n=h(C_n,D_n;\phi)$ from $C_n=\bigl(g_i(D_n;\theta_i)\bigr)_{i\in[K]}$, using the problem context and the (config, performance) trajectory $D_n$. \textbf{(3)} The selected configuration is evaluated on the real ML task, yielding the true performance $y_n=f(x_n)$. \textbf{(4)} The pair $(x_n,y_n)$ is appended, $D_{n+1}=D_n\cup\{(x_n,y_n)\}$, and consumed by both the tools and the Judge in later rounds. The Module~2 mechanisms introduced later change \emph{when} and \emph{how cheaply} each round runs; they alter the selected configuration only within the speculation tolerance $\tau$, and not at all on reject, which falls back to the identical serial selection.

\paragraph{Judge implementation: In-Context Learning Judge.}
We instantiate the Judge as a single \textbf{In-Context Learning (ICL) Judge} and deliberately keep it simple. 
We score each candidate in $C_n$ with a prompt that shares a common prefix (the problem context and the history $D_n$) and differs only in the queried candidate config; for each candidate we draw $S$ predictions in parallel and use their sample mean and standard deviation as the predictive mean and uncertainty, then select the candidate with the highest Expected-Improvement (EI) score. The sample spread is a practical uncertainty estimate rather than a calibrated posterior, so the decoding temperature acts as a direct explore/exploit knob. Framing selection as in-context regression plus EI, rather than free-form reasoning, keeps the Judge characterizable and gives it a clean acquisition geometry. Two properties of the LLM make it well suited to this role: 1) \textbf{prior-informed cold-start.} Because modern models and datasets are large, each evaluation is costly, so a realistic HPO budget spans only tens of evaluations (here $30$ to $50$), a regime in which classical surrogates are data-starved and cold. Having internalized hyperparameter regularities from pretraining (learning rates near $10^{-3}$ on a log scale, batch sizes as powers of two), the LLM yields informative predictions from the first rounds. 2) \textbf{conditioning on textual side information.} Alongside the $(x,y)$ history, the prompt supplies (i) task meta-information (model family, dataset size $n$, feature count $d$, the ratio $n/d$, task type), which grounds the pretraining prior in the concrete problem, so that, e.g., an XGBoost run on a small, high-dimensional dataset is steered toward shallower, more regularized trees than one on a large dataset; (ii) the search-space semantics of each hyperparameter (name, range, scale); and (iii) the optimization state in $D_n$ (round index, best-so-far, recent improvement), which supplies explore/exploit context. The ICL Judge is also cheap per round, a property the agent-system co-design in Module~2 is designed to exploit and analyzes in detail; the Judge prompt template is shown in Table~\ref{tab:judge-prompt}.

\section{ASAP Module 2: Agent-System Co-Design}

\subsection{Serial Workflow and Its Wall-Clock Cost}
\label{sec:method-overview}

Module~1 establishes \emph{which} configuration the integrated agent should pick each round: a diverse pool of tools proposes candidates and an LLM-as-Judge selects among them, beating the uninformed-selection baseline whenever the Judge outranks random selection over the candidate set (Eq.~\eqref{eq:condition}). That selection quality, however, comes at a price the classical HPO pipeline never pays. The Judge's ranking quality decides regret, but the LLM inference and tool calls it requires inflate the wall-clock of every round, and on long optimization runs this overhead can erode the very efficiency advantage an optimizer is meant to deliver. Module~1 fixes the per-round \emph{decision}; this section re-architects the loop so that the same decision is paid for \emph{at lower latency}, leaving the configuration trajectory (and hence the regret) unchanged within the speculation accept tolerance, and exactly unchanged whenever speculation is rejected.

To make the overhead precise, we decompose the wall-clock of the integrated loop. A standard LLM-coordinated HPO loop pays, each round, a tool-proposal cost, a Judge cost, and the cost of evaluating the selected configuration on the real ML task, so the total wall-clock over $N$ rounds is
\begin{equation}
T_{\text{wall}}^{\text{base}} = \sum_{n=1}^{N}\big(T_{\text{tool}}^{(n)} + T_{\text{judge}}^{(n)} + T_{\text{ML}}^{(n)}\big).
\label{eq:twall-base}
\end{equation}
A classical optimizer incurs a negligible proposal cost beside the evaluation term $T_{\text{ML}}^{(n)}$; the additional $T_{\text{tool}}^{(n)}+T_{\text{judge}}^{(n)}$ is precisely the overhead that makes an integrated LLM agent expensive in wall-clock despite its lower regret. Throughout this section, $y_n$ is the true performance at round $n$, $\hat y_n$ its prediction, $\tau$ the speculation accept threshold, $W$ the Self-Tuner period, and $g_i(\cdot)$ tool $i$'s proposal.

We attack the two overhead terms with three mechanisms. The tool-proposal term $T_{\text{tool}}$ is already collapsed to the slowest single tool by running the $K$ tools in parallel (Sec.~\ref{sec:integrated-loop}). The remaining two target the Judge: a \emph{KV-cache-aware prompt design} (Sec.~\ref{sec:c2-prefix}) that shrinks $T_{\text{judge}}$, and a \emph{cross-iteration speculation} scheme (Sec.~\ref{sec:c3-speculation}) that hides whatever tool$+$Judge cost remains by executing it in parallel with the dominant evaluation term $T_{\text{ML}}$. Together they drive $T_{\text{wall}}$ toward the classical lower bound $\sum_n T_{\text{ML}}^{(n)}$, altering the selected configuration only within the tolerance $\tau$ (and not at all on reject).

\subsection{Reducing $T_{judge}$: KV-Cache-Aware Prompts}
\label{sec:c2-prefix}

The Judge from Module~1 is invoked on the critical path of \emph{every} round, so its inference latency $T_{\text{judge}}$ is paid $N$ times over a run. Our key observation is that, across rounds, the Judge prompt is almost entirely repeated. Because we instantiate the Judge as an in-context regression task (Sec.~\ref{sec:integrated-loop}), each prompt is a fixed task description, followed by the historical $(\text{config},\text{performance})$ demonstrations in $D_n$, and closed by a short query asking for the predicted performance of one candidate. The history is \emph{append-only}: round $n{+}1$ differs from round $n$ only by the newly committed observation $(x_n,y_n)$ and the next query, while the entire leading block of tokens is token-for-token identical. This structural redundancy is exactly what a transformer KV cache can exploit: once a token's keys and values are computed, an identical prefix never needs to be prefilled again.

We therefore factor the Judge prompt into an \textbf{immutable prefix} and a \textbf{per-round suffix} (Table~\ref{tab:judge-prompt}). The prefix holds the task description together with the demonstrations committed so far; it is shared identically across the $K$ candidates scored within a round and grows append-only across rounds, so its KV cache is computed once and extended rather than recomputed. The suffix carries only the candidate config whose performance is to be predicted in the current round. The cache then grows incrementally: at round $n{+}1$, the just-committed pair $(x_n,y_n)$ is appended onto the existing cache so it joins the prefix going forward, without recomputing any earlier token.

\begin{table}[t]
\centering
\begin{tabular}{|p{0.96\columnwidth}|}
\hline
\rule{0pt}{2.4ex}\textbf{Immutable prefix} \;\textit{(task description $+$ committed demonstrations through round $n{-}1$; prefilled once; KV cache reused across rounds and candidates)} \\[2pt]
{\footnotesize\ttfamily The following are hyperparameter configurations for a XGBoost and the corresponding performance measured in accuracy. The model is evaluated on a tabular classification task and the label contains 3 classes. The tabular dataset contains 27025 samples and 6 features (0 categorical, 6 numerical). Your response should only contain the predicted accuracy in the format \#\# performance \#\#.}\\[3pt]
{\footnotesize\ttfamily \textit{-- observed history ($D_n$): config $\rightarrow$ true accuracy --}}\\[1pt]
{\footnotesize\ttfamily Hyperparameter configuration: colsample\_bytree is 0.6, eta is 0.0500823341, max\_depth is 4, reg\_lambda is 0.5445010154 \quad Performance: \#\# 0.867788 \#\#}\\[1pt]
{\footnotesize\ttfamily Hyperparameter configuration: colsample\_bytree is 0.6, eta is 0.2360640703, max\_depth is 46, reg\_lambda is 0.0382369760 \quad Performance: \#\# 0.930664 \#\#}\\[1pt]
{\footnotesize\ttfamily \dots\ \;(remaining committed demonstrations in $D_n$)}\\[1pt]
{\footnotesize\ttfamily Hyperparameter configuration: colsample\_bytree is 0.5, eta is 0.0015978643, max\_depth is 1, reg\_lambda is 0.0012671438 \quad Performance: \#\# 0.703050 \#\#}\\
\hline
\rule{0pt}{2.4ex}\textbf{Round $n$ suffix} \;\textit{(the query candidate; only this is prefilled at round $n$)} \\[2pt]
{\footnotesize\ttfamily Hyperparameter configuration: colsample\_bytree is 0.4, eta is 0.5069423632, max\_depth is 20, reg\_lambda is 2.0054749362 \quad Performance: \#\#\,$\rightarrow$ predict\,\#\#}\\
\hline
\end{tabular}
\caption{KV-cache-aware Judge prompt. A long immutable prefix (task description $+$ committed $(\text{config},\text{performance})$ demonstrations up through round $n{-}1$) is prefilled once and its KV cache is reused; round $n$ only appends a short suffix (the new candidate query), so only the suffix is prefilled. After round $n$ commits, $(x_n,y_n)$ extends the prefix for round $n{+}1$ without recomputing earlier tokens.}
\label{tab:judge-prompt}
\end{table}

With KV-cache reuse, per-round prefill cost drops from the full prompt length to the suffix length,
\begin{equation}
T_{\text{prefill}}^{\text{cached}} \;\approx\; L_{\text{suffix}}\,t_{\text{token}}
\;\;\ll\;\;
L_{\text{full}}\,t_{\text{token}} = T_{\text{prefill}}^{\text{full}},
\label{eq:prefill-cached}
\end{equation}
where $t_{\text{token}}$ is the per-token prefill cost. As long as the prefix is reused, only suffix-length prefill is paid each round, shrinking the \emph{prefill} term of $T_{\text{judge}}$ from $O(L_{\text{full}})$ to $O(L_{\text{suffix}})$; since the Judge decodes only a few performance tokens, prefill dominates $T_{\text{judge}}$, so this is the dominant per-round saving.

\subsection{Hiding Overhead via Cross-Iteration Speculation}
\label{sec:c3-speculation}

\paragraph{Motivation.}
As analyzed in Sec.~\ref{sec:method-overview}, the three stages of each round (tool proposal, LLM-as-Judge, and ML evaluation) run strictly serially because each consumes the output of the previous one. The ML evaluation $T_{\text{ML}}$ typically dominates, yet throughout this long stage the tools and the Judge sit idle, waiting only for the single scalar the evaluation returns. Our design exploits two observations. First, this scalar need not be waited for: when the Judge selects $x_n$, it has already predicted that config's performance $\hat\mu(x_n)$ as part of its EI scoring (Sec.~\ref{sec:integrated-loop}), so an estimate of the outcome $y_n=f(x_n)$ is available for free before the evaluation even starts. Second, speculative execution (predicting a not-yet-available value and proceeding on the prediction) is a proven way to remove serial dependencies and accelerate agentic pipelines. We combine the two: rather than wait for round $n$'s true performance $y_n$, we proceed on the Judge's already-computed prediction of $x_n$ and run the next round's tools and Judge ahead of time, in parallel with the in-flight evaluation, so their cost is hidden under $T_{\text{ML}}^{(n)}$ instead of paid after it. We call this \emph{cross-iteration} speculation because the speculative work belongs to round $n{+}1$ while the quantity being predicted ($y_n$) is produced by round $n$: the speculation bridges two consecutive optimization iterations rather than overlapping sub-steps within one.

\paragraph{(a) Reusing the Judge's prediction.}
We take the Judge's own predicted performance of the selected config as the speculated outcome: $\hat y_n \;\triangleq\; \hat\mu(x_n)$, where $\hat\mu(x_n)$ is the sample-mean prediction the Judge already computed when scoring $x_n$ (Sec.~\ref{sec:integrated-loop}); no separate predictor is introduced, and the estimate conditions on $x_n$ and the history $D_n$, exactly the inputs the Judge used to select it. Writing $\hat D_{n+1}=D_n\cup\{(x_n,\hat y_n)\}$ for the speculative history, the system executes round $n{+}1$ on $\hat D_{n+1}$ while round $n$'s evaluation is still running: the tools produce proposals $g_i(\hat D_{n+1};\theta_i)$ and the Judge selects a speculative configuration $h(\hat C_{n+1},\hat D_{n+1};\phi)$, all overlapped with the running evaluation. Speculation is one step deep, each round predicting a single not-yet-returned $y_n$ from committed history, so no prediction is ever conditioned on another prediction.

\paragraph{(b) Accept test.}
Once the real $y_n$ returns, we measure the prediction's relative error and accept the speculative round only when it is within the current tolerance $\tau$:
\begin{equation}
a_n \;=\; \mathbb{1}\!\left[\,\frac{|\hat{y}_n - y_n|}{|y_n|} < \tau\,\right].
\label{eq:accept-test}
\end{equation}
On accept ($a_n=1$), the speculatively chosen configuration is dispatched to ML evaluation immediately, and the tool$+$Judge cost (already paid in parallel with $T_{\text{ML}}^{(n)}$) is fully hidden. On reject ($a_n=0$), we re-run only the tool$+$Judge stage on the true $y_n$, which is exactly the non-speculative baseline for that round; the dominant ML-evaluation stage is never re-run for verification. The per-round worst case therefore equals the serial baseline: on the latency axis, speculation can only save wall-clock, never lose it. When round $n{+}1$'s tool$+$Judge work is overlapped with round $n$'s evaluation, the per-round wall-clock collapses to
\begin{equation}
T_{\text{round}}^{(n)} \;=\; \max\!\big(T_{\text{ML}}^{(n)},\; T_{\text{tool}}^{(n+1)}+T_{\text{judge}}^{(n+1)}\big),
\label{eq:t-round}
\end{equation}
so once the overhead fits under $T_{\text{ML}}^{(n)}$ it is fully hidden and $T_{\text{wall}}$ approaches the lower bound $\sum_n T_{\text{ML}}^{(n)}$. The threshold $\tau$ is thus a wall-clock/regret knob: as $\tau\!\to\!0$ the speculative path reduces to the serial agentic workflow. We do not hand-tune it, but let the Self-Tuner (Sec.~\ref{sec:c4-self-evolver}) adapt it from the realized prediction errors and downstream gain, keeping regret the primary objective and wall-clock the secondary one.

\subsection{LLM as Self-Tuner: Adapting the Speculation Threshold from Experience}
\label{sec:c4-self-evolver}

The speculation accept test in Sec.~\ref{sec:c3-speculation} hinges on a single knob, the threshold $\tau$, which cannot be fixed once and for all: it is task-dependent and best learned online, since too loose a $\tau$ admits inaccurate speculations that hurt regret, while too tight a $\tau$ rejects often and recovers little latency. We therefore cast $\tau$ as an adaptive decision managed by a \emph{Self-Tuner}: an off-critical-path background process invoked every 5 rounds over a sliding window of the most recent 5 rounds. Because it runs in parallel with the main loop and is allowed to be slow, its latency never enters $T_{\text{round}}$, so it can use a capable LLM to reason over the log. It reads execution logs of the form $\{(\hat y_m, y_m, a_m, \Delta_m)\}$, where $\Delta_m$ is the downstream gain of round $m$ (the improvement in best-so-far it produced), and updates $\tau$ as follows: The Self-Tuner sets $\tau$ to a quantile of the recently observed relative prediction errors,
\begin{equation}
\tau \;\leftarrow\; Q_{q}\!\Big(\big\{\, r_m \,\big\}_{m=n-W+1}^{\,n}\Big),
\qquad
r_m \;=\; \frac{|\hat y_m - y_m|}{|y_m|},
\label{eq:tau-quantile}
\end{equation}
and uses the LLM to choose the quantile level $q$ for the next window by weighing the previous window's realized accept rate against its downstream gain. Reasoning over the raw log, rather than tracking a fixed accept-rate target, lets it react to regime shifts (a drift in the error distribution, or a transition from exploration to exploitation) that a static rule would miss.
Overall, the Self-Tuner replaces a hand-pinned $\tau$ with an online process that adapts the threshold from the task's own trajectory, and is the only online-tuned component in ASAP; its updates take effect at the start of the next window. Consistent with the rest of the framework, it follows a lexicographic priority (regret first, wall-clock second), adjusting $\tau$ to reduce wall-clock while aiming to leave optimization quality intact.

\section{Experiments}

\label{Sec:exp}
We organize our evaluation around the design of ASAP into three research questions, each analyzed from multiple perspectives:

\begin{itemize}[leftmargin=*,itemsep=2pt,parsep=0pt]
    \item \textbf{RQ1 (Effectiveness and Generalization).} Does ASAP reach lower regret per unit wall-clock?
    \item \textbf{RQ2 (Decomposing the wall-clock advantage).} How does each component contribute to ASAP's regret-per-wall-clock?
    \item \textbf{RQ3 (Interpretability and analysis).} Why does ASAP work and further in-depth analysis?
\end{itemize}

\subsection{Experiment Settings}
\paragraph{Benchmarks.}
Since ASAP targets HPO settings where per-evaluation cost is non-trivial, we evaluate on two complementary benchmark families that jointly cover the regimes practical HPO faces, with deliberate emphasis on modern, realistic workloads.
(1) \textbf{HPOBench}~\cite{eggensperger2022hpobenchcollectionreproduciblemultifidelity}, real training of XGBoost and/or RandomForest across six OpenML tabular tasks (\textit{jasmine}, \textit{adult}, \textit{madeline}, \textit{volkert}, and two larger high-dimensional tasks), for 9 (task, model) combinations spanning binary, multi-class (up to 100-way), and high-dimensional regimes (3\,k--58\,k samples $\times$ 6--1646 features), with measured per-trial wall from $\sim$9\,s to $\sim$240\,s.
(2) \textbf{PD1}~\cite{wang2024pretrainedgaussianprocessesbayesian}, Google HyperBO's pre-computed deep-learning HPO replay spanning vision (e.g.\ Wide-ResNet on CIFAR-10/100/SVHN, ResNet-50 on ImageNet, Simple-CNN/Max-Pooling-CNN on MNIST/Fashion-MNIST), language (Transformer on LM1B and WMT En--De), and protein (Transformer on Uniref50), for 19 (task, model) combinations, with per-trial GPU training wall from $\sim$2\,min to $\sim$48\,h.
In total the evaluation covers \textbf{28 (task, model) combinations} (9 HPOBench $+$ 19 PD1), seven model architectures spanning classical tabular learners, modern CNNs, and modern Transformers, with per-trial wall covering four orders of magnitude ($\sim$9\,s to $\sim$172\,k\,s), all within the non-trivial-eval regime ASAP targets. The full breakdown (dataset shapes, search-space dimensionality, per-trial ML eval time, and objectives) appears in Appendix~\ref{sec:bench-details} (Table~\ref{tab:bench-tasks}).

\paragraph{Baselines.}
We compare against two families of baselines.
(1) \textit{Statistical baselines}: GP, TPE, Optuna, SKOPT, TuRBO, SMAC, DNGO.
(2) \textit{LLM-based baseline}: LLAMBO, which uses an LLM to propose hyperparameters one at a time rather than coordinate a pool of proposers.

\paragraph{Our method (two variants).}
ASAP itself is evaluated as two variants of our own method: \textbf{ASAP}, the full system with cross-iteration speculation and the Self-Tuner (Module~1 $+$ Module~2), and \textbf{ASAP$^{-\text{spec}}$}, the same fully-integrated Module~1 but with cross-iteration speculation and the Self-Tuner disabled (the ``no speculate'' line in the figures). Reporting both makes the Module-1 vs.\ Module-2 contributions directly visible.

\begin{figure*}[tp]
    \centering
    \begin{minipage}{0.62\textwidth}
        \centering
        \includegraphics[width=\textwidth]{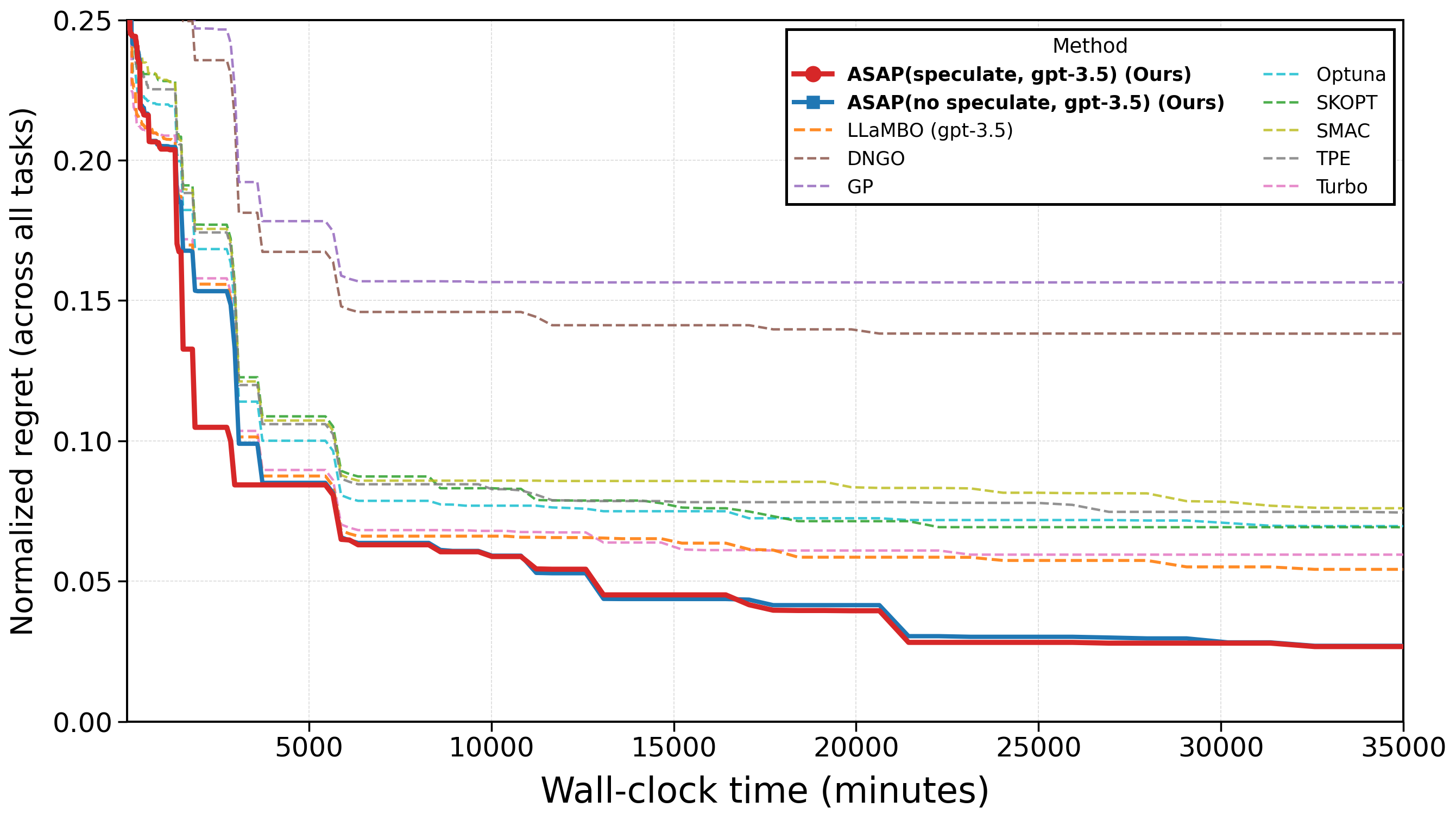}
        \\\textbf{(a) Regret vs.\ end-to-end wall-clock (primary axis)}
    \end{minipage}\hfill
    \begin{minipage}{0.36\textwidth}
        \centering
        \includegraphics[width=\textwidth]{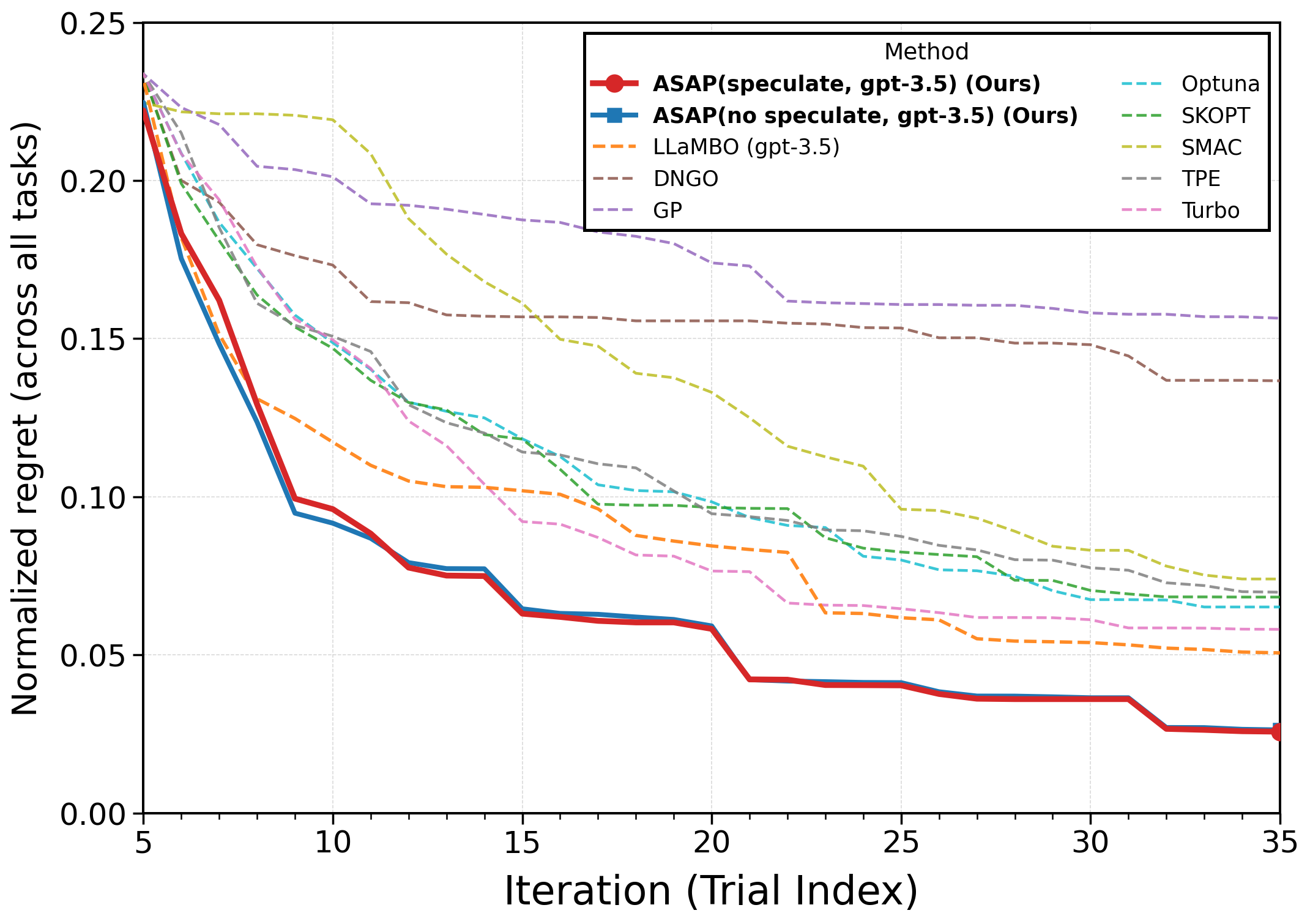}
        \\\textbf{(b) Regret vs.\ iteration (auxiliary)}
    \end{minipage}
    \caption{Normalized regret vs.\ (a) end-to-end wall-clock and (b) iteration, aggregated over the 28 combinations from HPOBench and PD1. Both \textbf{ASAP (speculate, ours; red)} and \textbf{ASAP (no speculate, ours; blue)} reach the lowest regret and nearly coincide on both axes, since speculation removes propose$+$judge overhead without impairing regret.}
    \label{fig:regret-vs-wallclock}
\end{figure*}

\begin{figure}[h]
    \centering
    \includegraphics[width=0.85\columnwidth]{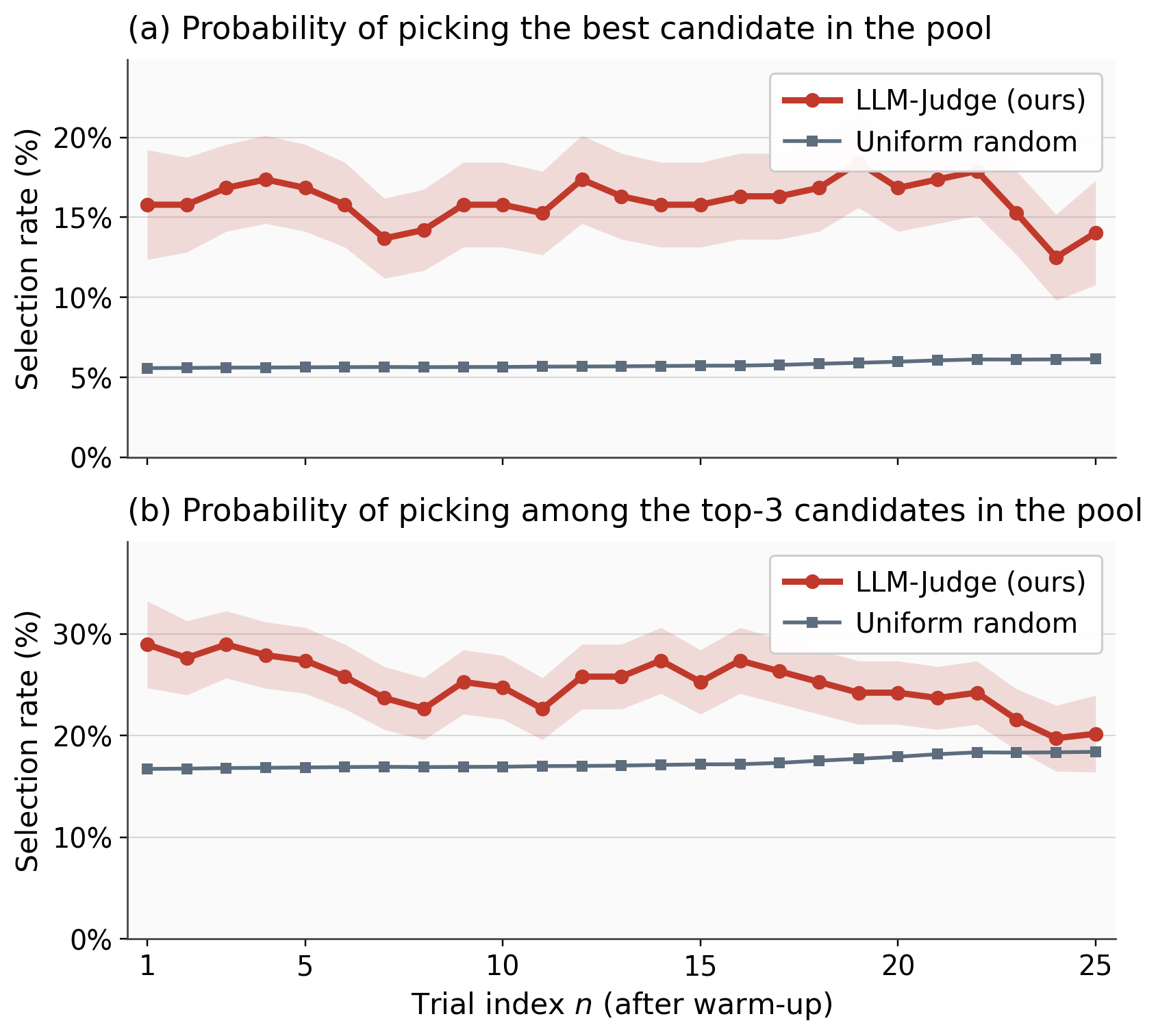}
    \caption{Per-trial pick-best and top-3 rates of the LLM Judge (red) vs.\ uniform-random selection over $C_n$ (grey; 5-trial centered moving average, three seeds). The Judge stays above random at \emph{every} trial for both metrics, especially from the first post-warm-up iteration onward.}
    \label{fig:judge-vs-random}
\end{figure}

\begin{figure}[h]
    \centering
    \includegraphics[width=0.9\columnwidth]{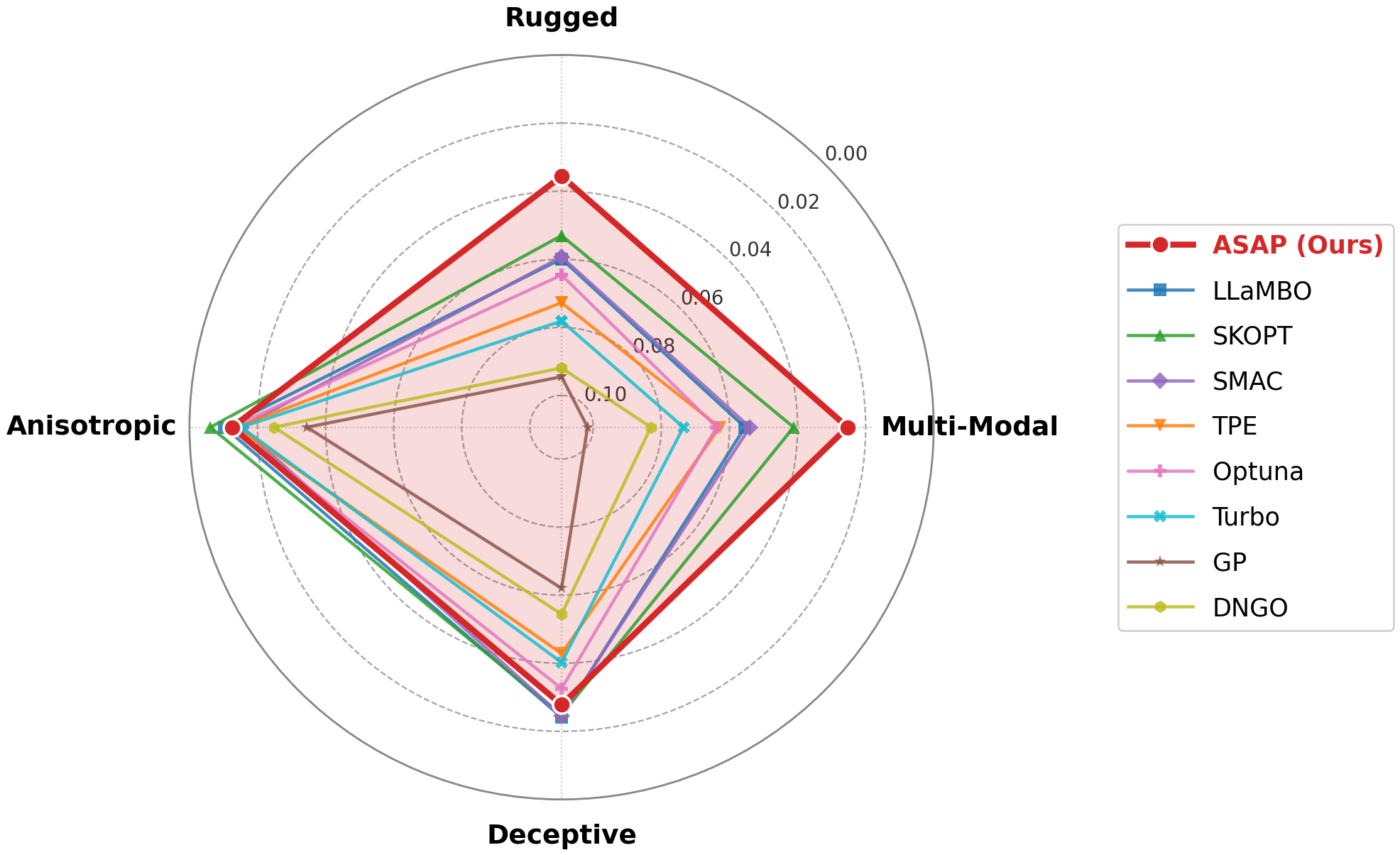}
    \caption{Performance by optimization-landscapes. \textbf{ASAP (Ours; red)} hugs the outer ring on every axis, while every single-bias BO tool drops on at least one: integration turns diversity into uniform robustness.}
    \label{fig:landscape-radar}
\end{figure}

\paragraph{Setup.}
We disable warm-starting and initialize all methods with the same five randomly sampled points per run. Each method runs three independent seeds, with 5 initial $+$ 30 iteration trials per task per seed (35 evaluations $\times$ 28 task-models $\times$ 10 methods $\times$ 3 seeds); reported curves average the three seeds and shaded bands denote $\pm1$ standard deviation. Hyperparameter search spaces appear in Appendix~\ref{sec:hyper_space}. For LLAMBO and all ASAP runs we use GPT-3.5-turbo-0125 as the LLM (the proposer in LLAMBO; the Judge and the Self-Tuner in ASAP), matching the LLAMBO baseline so the comparison is on the integration mechanism rather than the underlying language model; a stronger LLM would likely improve all LLM-based methods further. All experiments run on identical CPU-only nodes (dual-socket AMD EPYC 7543 32-core processors, 64 cores per node, $\sim$250\,GB RAM) so that the wall-clock comparison across methods is unaffected by hardware variation; LLM inference is served by the OpenAI API, so the same compute environment carries every method's full propose-evaluate loop. For PD1 we add the precomputed per-trial GPU training time to every method's measured wall-clock so the comparison reflects what an end-to-end deployment would actually pay.

\paragraph{Evaluation Metrics.}
For classification we use accuracy, for regression negative MSE. For both we report normalized regret at trial $t$,
\begin{equation}
\text{Regret}_t \;=\; \frac{s^*_{\max} - \max_{h\in\mathcal H_t} f(h)}{s^*_{\max} - s^*_{\min}},
\end{equation}
where $\mathcal H_t$ is the set of configurations evaluated up to trial $t$, $f(h)$ is the (higher-is-better) performance of $h$, and $s^*_{\max}, s^*_{\min}$ are the best and worst scores observed for that task across all methods and trials. Regret lies in $[0,1]$, equals $0$ once the best seen configuration is found, and is lower-is-better. Our \textbf{primary metric is regret as a function of end-to-end wall-clock}, the lowest normalized regret reached per unit of real time, which jointly captures decision quality and the overhead of producing each decision and is the axis prior iteration-centric HPO ignores. We report per-iteration regret only as an \emph{auxiliary} axis: it isolates decision quality from timing and keeps ASAP comparable to prior iteration-based HPO.

\subsection{RQ1: Effectiveness and Generalization}
\label{sec:rq1-eff}

ASAP's headline result is how low a regret it reaches per unit of end-to-end wall-clock. We report regret vs.\ wall-clock against all baselines (Fig.~\ref{fig:regret-vs-wallclock}), with per-iteration regret as the auxiliary axis.

\paragraph{ASAP dominates every baseline, and speculation removes overhead where it is non-negligible.}
\Result{Both ASAP variants reach a far lower regret floor ($\sim$0.026) than every baseline ($\sim$0.05--0.16) on both axes; speculation cuts propose$+$judge latency by up to 25.6\% on a single HPOBench combo, with no loss in per-iteration sample efficiency.} The dominant effect (Fig.~\ref{fig:regret-vs-wallclock}) is the integration win: on both axes, ASAP and ASAP$^{-\text{spec}}$ reach $\sim$0.026 regret, vs.\ $\sim$0.05 for the best baseline (LLAMBO), $\sim$0.058 for the best statistical tool (TuRBO), and 0.07--0.16 for the rest, roughly $2\text{--}6\times$ higher. The two variants overlap on the iteration axis (b), so speculation leaves the trajectory essentially unchanged; they nearly coincide on wall-clock too, because PD1's multi-hour $T_{\text{ML}}$ dwarfs the propose$+$judge overhead and dominates the 28-task aggregate. The benefit thus concentrates where overhead is non-negligible: on HPOBench, where $T_{\text{propose}}$ and $T_{\text{ML}}$ are comparable, hiding propose$+$judge under the evaluation saves up to \textbf{25.6\% per-trial on the best combo} and \textbf{$\sim$10\% across the nine combos}; on PD1, $T_{\text{propose}}{:}T_{\text{ML}}\approx 10:10^{3}\text{--}10^{5}$ leaves only a few percent, the regime where overhead was never the bottleneck. The overlap in (b) confirms this saving costs no sample efficiency.

\paragraph{Whether and why the LLM Judge beats random selection.}
\Result{The LLM Judge picks the best candidate in the pool about
$3\times$ more often than uniform-random selection, and does so from the very first
iteration after warm-up.}
Sec.~\ref{sec:integrate-vs-replace} established Eq.~\eqref{eq:condition}
as a \emph{per-round, shared-history} precondition: on a given committed history
$D_n$, the integrated agent improves on uniform selection over $C_n$ exactly when the
Judge outranks random over that same set. Here we verify that this precondition holds
empirically. At every iteration of an ASAP run we record $C_n$, look up the
ground-truth $f(x)$ for each candidate, and compare the Judge's pick against
uniform-random selection over $C_n$. The Judge has genuine selection ability: its
pick-best rate ($\sim$15--17\%) is about $3\times$ the random baseline
$1/|C_n|\approx 5.6\%$, and its top-3 rate is $\sim$1.5--1.7$\times$ the $3/|C_n|$
baseline, so Eq.~\eqref{eq:condition} is satisfied at every trial $n\in[1,25]$
(Fig.~\ref{fig:judge-vs-random}). The advantage is already present at $n\approx 1$
and stays flat rather than rising with data---direct evidence for the prior-informed
cold-start of Sec.~\ref{sec:integrated-loop}: with almost no trajectory to fit at
$n\approx 1$, the ability must come from the LLM's pretraining prior over
hyperparameters, not from regression on a still-empty history. RQ1 then shows this
per-round advantage carries through to the lowest end-to-end regret of any method;
the per-round-to-trajectory link is empirical, as noted in
Sec.~\ref{sec:integrate-vs-replace}.

\paragraph{Robust performance across optimization landscapes.}
\Result{ASAP stays in the top group on all four landscape axes, whereas each single-bias optimizer collapses on at least one; integration converts tool diversity into uniform robustness.}
The structural-heterogeneity argument of Sec.~\ref{sec:integrate-vs-replace} (each single-bias tool excels only where its prior matches the landscape) predicts that no fixed tool is robust across landscapes and that integration removes this fragility. We classify the 28 tasks along four flacco-style~\cite{kerschke2017comprehensivefeaturebasedlandscapeanalysis} axes (Rugged, Multi-Modal, Deceptive, Anisotropic), computed with pflacco from the (config, value) points every method already evaluated (no extra runs; definitions in Appendix~\ref{sec:landscapes}). Per axis we keep the top half of tasks and compare mean normalized regret at iteration 35 (Fig.~\ref{fig:landscape-radar}; radial axis inverted, outer $=$ lower regret). ASAP hugs the outer ring on every axis: strongest on Rugged and Multi-Modal, and in the top group on Deceptive and Anisotropic, whereas every single-bias BO tool drops on at least one (e.g.\ GP and DNGO exceed 0.09 on Rugged, where their smooth-kernel/network prior mismatches the jagged landscape). Integration turns the diversity that defeats single tools into uniform robustness.

\begin{figure}[t]
    \centering
    \includegraphics[width=0.6\columnwidth]{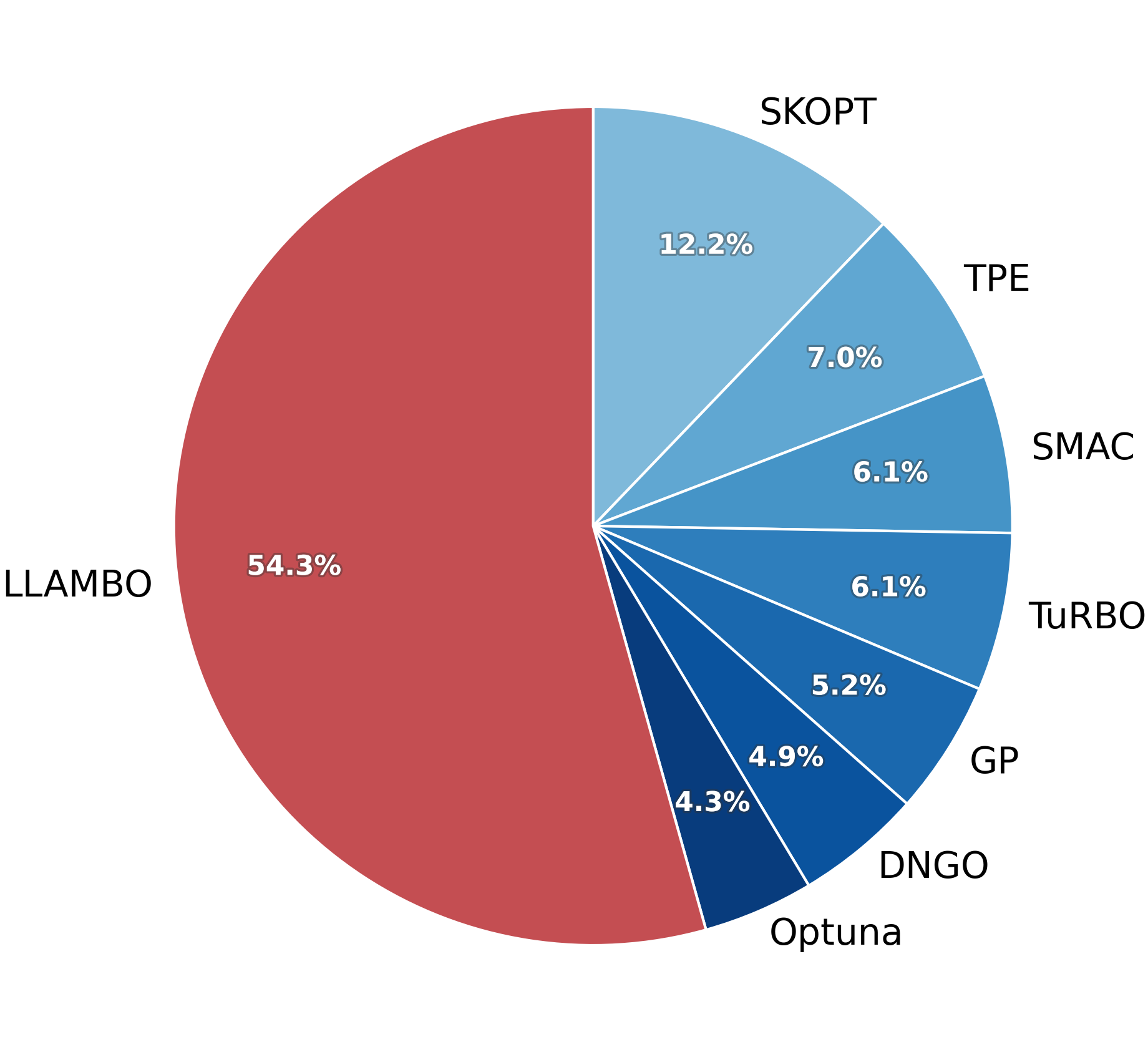}
    \caption{Fraction of Judge selections per tool.}
    \label{fig:exp_unbiased}
\end{figure}

\begin{figure}[h]
    \centering
    \includegraphics[width=\columnwidth]{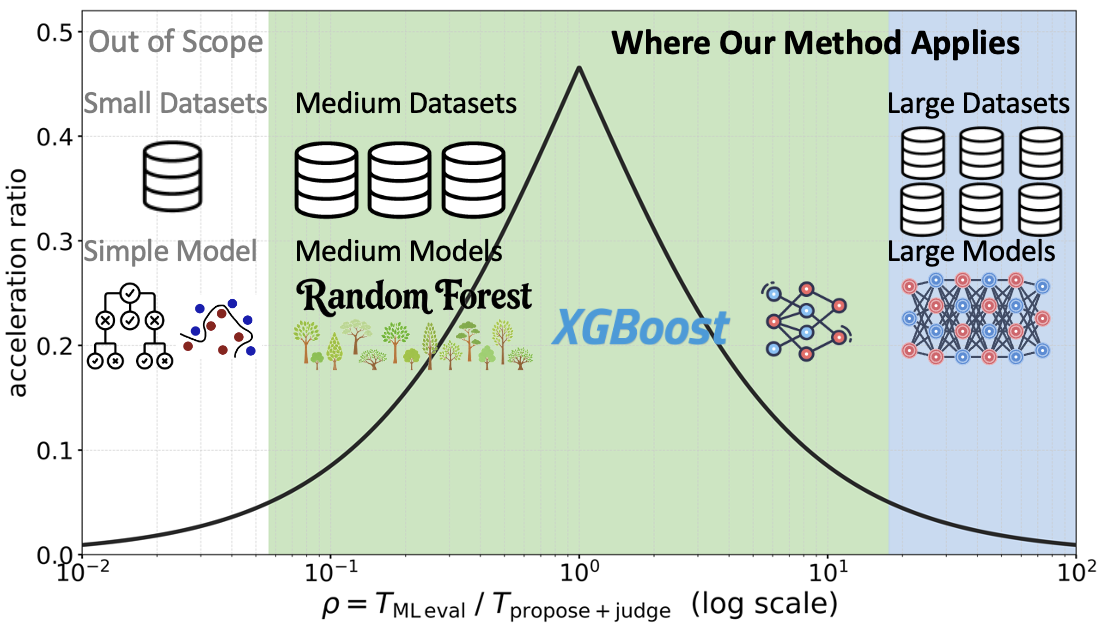}
\caption{Where ASAP applies and where speculation helps, as a function of the eval-to-overhead ratio $\rho = T_{\text{ML}}/(T_{\text{tool}}+T_{\text{judge}})$ (log scale). Speculation's acceleration peaks at $\rho\!\approx\!1$ (medium models such as RandomForest and XGBoost), tapers for very large models ($\rho\!\gg\!1$) where ASAP still wins on regret through integration, and is out of scope only for trivial-eval tasks ($\rho\!\ll\!1$).}
    \label{fig:spec-ratio}
\end{figure}

\subsection{RQ2: Decomposing the Wall-Clock Gain}
\label{sec:rq2-efficiency}

\begin{figure}[h]
    \centering
    \includegraphics[width=0.85\columnwidth]{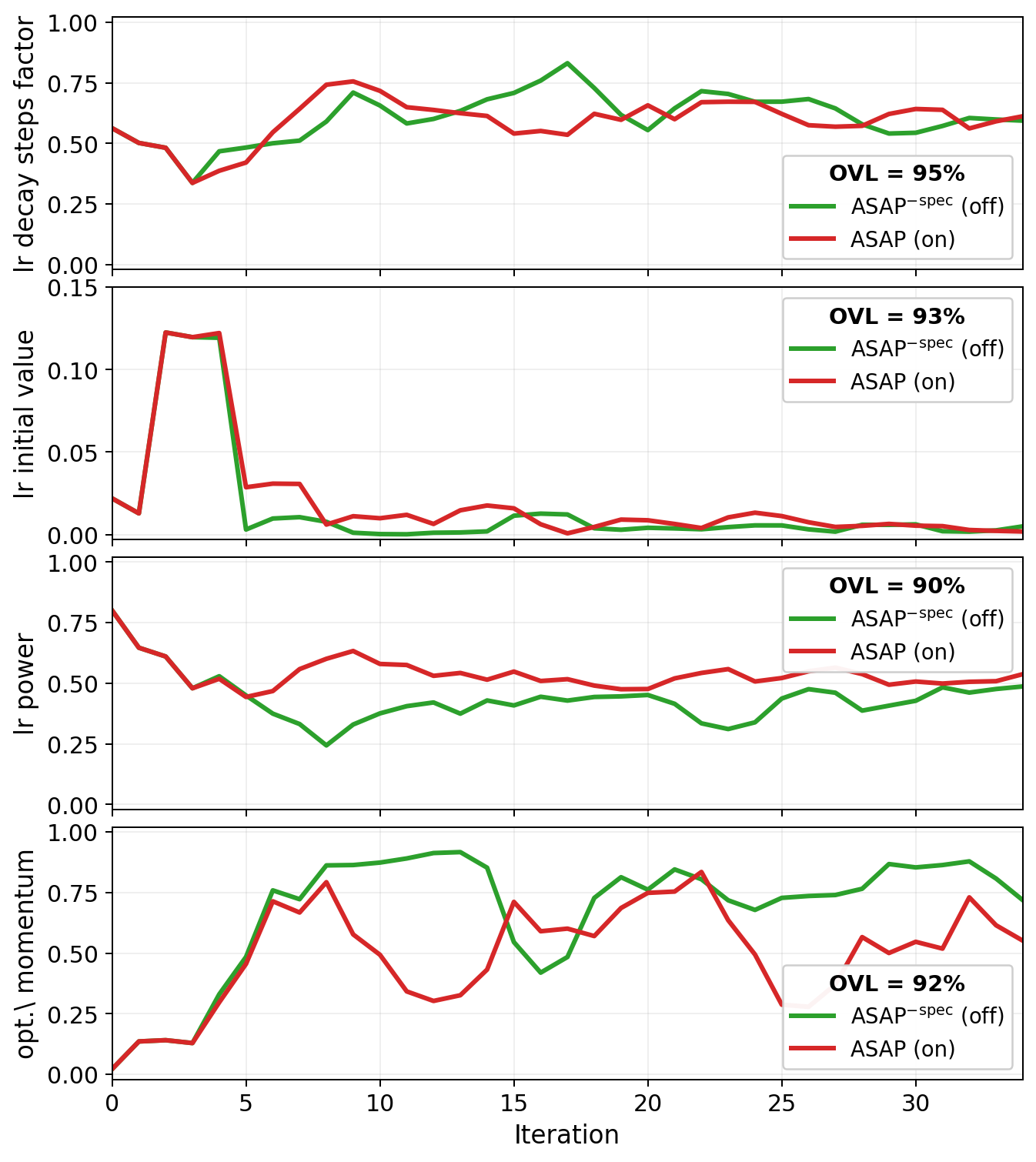}
    \caption{Speculation does not change which hyperparameters are explored. Per-round median of each evaluated hyperparameter (normalized to $[0,1]$) for ASAP (speculation on, red) vs.\ ASAP$^{-\text{spec}}$ (off, green). The two track each other throughout, pooled per-hyperparameter distributions overlapping by $90$--$95\%$ (OVL, per legend): speculation reorders only \emph{when} the tool$+$Judge stage runs, not \emph{where} the agent searches.}
    \label{fig:spec-hp-agreement}
\end{figure}

\begin{figure}[h]
    \centering
    \includegraphics[width=0.95\columnwidth]{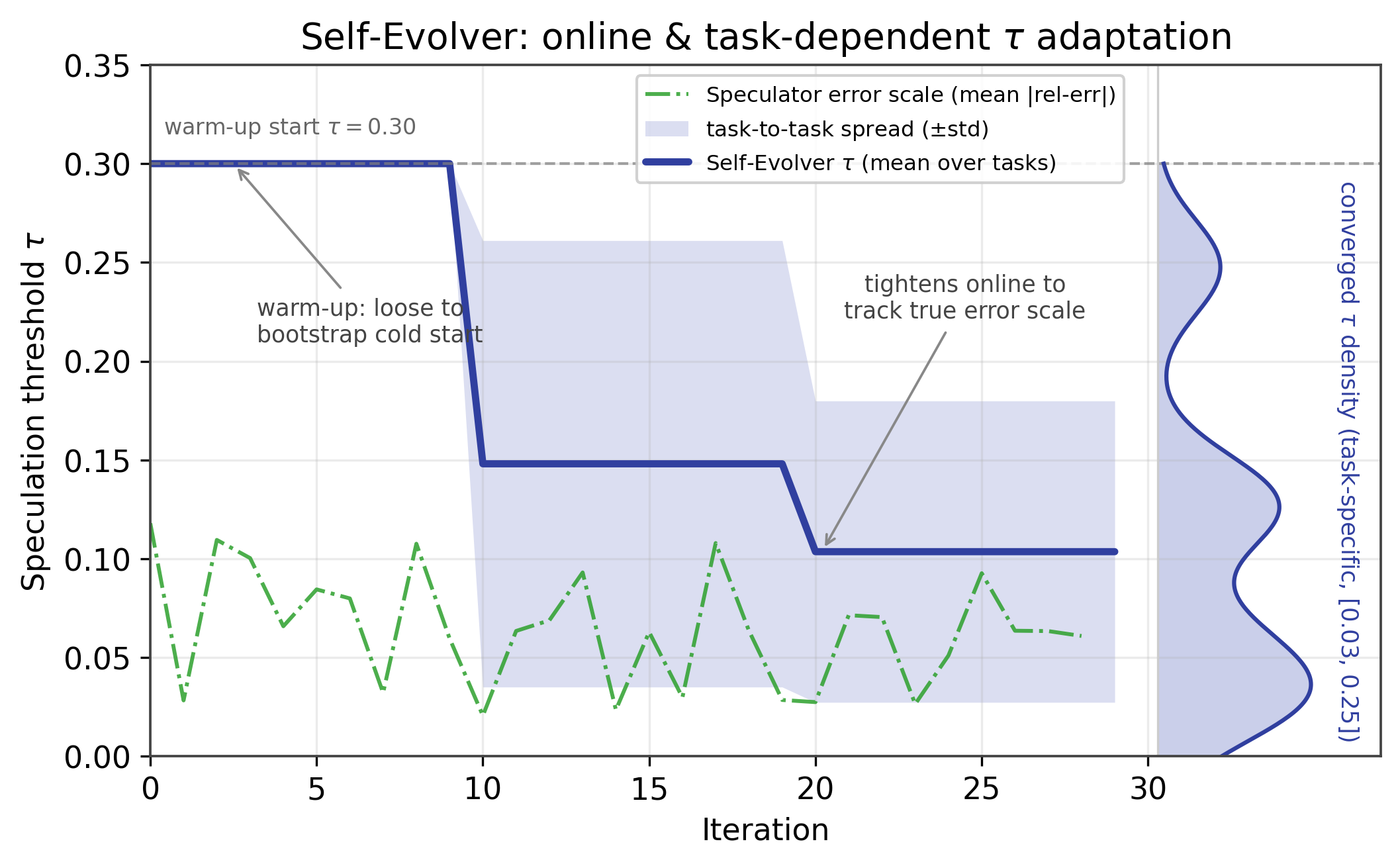}
    \caption{Self-Tuner behavior on the speculate-effective HPOBench tasks. The adaptive threshold $\tau$ (indigo, mean over tasks) starts loose at the warm-up $0.30$ and tightens online to ${\sim}0.10$, tracking the empirical prediction-error scale (green, mean $|\text{rel-err}|$); the $\pm$std band and right-margin density show convergence to task-specific values spanning $[0.03,0.25]$.}
    \label{fig:self-evolver}
\end{figure}

\begin{figure*}[t]
    \centering
    \includegraphics[width=\textwidth]{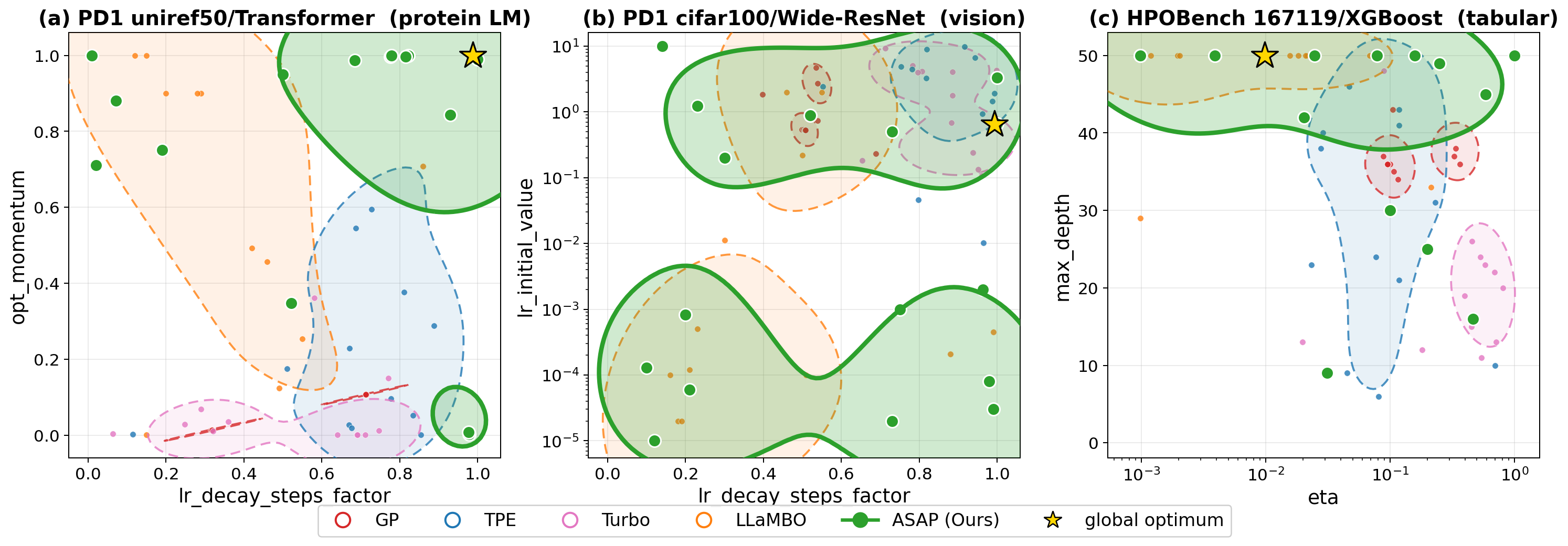}
    \caption{Case studies (RQ3) on three (benchmark, model) combinations (PD1/Transformer, PD1/Wide-ResNet, HPOBench/XGBoost), spanning single-proposer failure modes: no single proposer reaches the optimum (a, b), or one grazes it without converging (c). Dashed blobs are each proposer's KDE contour (GP, TPE, TuRBO, LLAMBO) with a sparse query sample; the green dots are ASAP's queries. Only ASAP's queries reach the gold-star optimum, driven by the Judge switching between proposers across iterations.}
    \label{fig:case-study}
\end{figure*}

RQ1 established ASAP's regret-per-wall-clock dominance; here we attribute it to its components. The two modules act on the regret-vs-wall-clock curve through complementary routes: Module~1 (integration) sets how low regret can go (the curve's height, kept by every ablation below), while Module~2, layered on the same agent, compresses the wall-clock to reach it (its horizontal position). We ablate each component and read both regret and wall-clock: removing integration collapses the regret floor; removing a Module-2 mechanism inflates wall-clock at the same regret. We then characterize the speculation and Self-Tuner behavior driving the latter.

\paragraph{Ablation: contribution of each component.}
\Result{Module~1 (integration) owns the regret floor and Module~2 (speculation) owns the wall-clock saving: removing integration nearly doubles regret, while removing speculation raises wall-clock at unchanged regret.}
We ablate each component and report per-task regret and end-to-end wall-clock, isolating its effect on regret-per-wall-clock. \textbf{ASAP$^{-\text{int}}$} ($\equiv$ LLAMBO) replaces the integrated pool with a single LLM-as-proposer, exposing the regret floor the Module-2 speedups operate above; \textbf{ASAP$^{-\text{spec}}$} disables cross-iteration speculation (tools and Judge run serially after each $T_{\text{ML}}$). In Table~\ref{tab:rq2-ablation}, removing integration raises regret (Module~1's quality), while removing speculation raises wall-clock at unchanged regret.

\begin{table}[h]
\centering
\resizebox{0.49\textwidth}{!}{%
\begin{tabular}{lcccc}
\hline
Variant & Avg.\ Regret $\downarrow$ & Wall (HPOBench, s) $\downarrow$ & Wall (PD1, s) $\downarrow$ & Speedup vs.\ vanilla$^\ddagger$ $\uparrow$ \\ \hline
\textbf{ASAP (full)}                & \textbf{0.026} & 18{,}063 & 81{,}884 & 1.13$\times$ \\
ASAP$^{-\text{int}}$ ($\equiv$ LLAMBO) & 0.051 & 13{,}656 & 82{,}637 & 1.50$\times$ \\
ASAP$^{-\text{spec}}$ (no speculation)  & 0.026 & 20{,}422 & 82{,}018 & 1.00$\times$ \\ \hline
\end{tabular}}
\caption{Ablation study: final regret and end-to-end wall-clock per family (median over 9 HPOBench / 19 PD1 tasks at the 30-iteration budget). $^\ddagger$Speedup vs.\ vanilla is HPOBench wall relative to the serial ASAP$^{-\text{spec}}$ baseline ($>\!1\times$ = faster).}
\label{tab:rq2-ablation}
\end{table}

\paragraph{The integrated pool is genuinely used.}
\Result{Commits are split across both sources: the LLM proposer
contributes a substantial share, and the seven BO tools jointly contribute a
comparable share, with no single BO tool dominating, so ASAP behaves as a genuine
mixture rather than collapsing onto any one proposer.}
Fig.~\ref{fig:exp_unbiased} reports the fraction of committed picks per tool from
exact per-round attribution. The LLM proposer (LLAMBO) accounts for 54.3\% of
commits, while the seven BO tools together account for the remaining 45.8\%. Neither
source is idle: roughly half of all committed configurations originate from the
classical BO pool, and within that pool the contribution is spread out rather than
concentrated (SKOPT 12.2\%, TPE 7.0\%, SMAC and TuRBO 6.1\% each, GP 5.2\%, DNGO
4.9\%, and Optuna 4.3\%). The Judge therefore draws on the whole pool across rounds:
the LLM proposer and the diverse BO tools each supply a meaningful fraction of the
final trajectory, which is exactly the complementary-bias integration ASAP is
designed to exploit (Sec.~\ref{sec:integrate-vs-replace}), not a wrapper around a single
proposer.

\paragraph{Speculation behavior.}
\Result{Speculation hides the propose+Judge latency under the ML evaluation while leaving the searched configurations essentially unchanged, so the wall-clock it saves comes at no regret cost.}
Against ASAP$^{-\text{spec}}$, ASAP reaches essentially the same final regret, and the two pick the same config most rounds (the speculative pick on $\hat D_{n+1}$ agrees with the true-history pick on $D_{n+1}$). The saved wall-clock (Table~\ref{tab:rq2-ablation}) is therefore free: a rejected round re-runs only tool$+$Judge and never re-invokes $T_{\text{ML}}$ (Sec.~\ref{sec:c3-speculation}), so the worst case equals the serial baseline. Agreement holds on which configurations are explored: Fig.~\ref{fig:spec-hp-agreement} pools the per-hyperparameter distribution of evaluated configs. The median trajectories of ASAP and ASAP$^{-\text{spec}}$ track each other on all four across the whole run, and their pooled distributions overlap by $90$--$95\%$ (overlapping coefficient, OVL). Speculation thus reorders only \emph{when} the tool$+$Judge stage runs (hiding it under $T_{\text{ML}}$), not \emph{where} the agent searches, which is why the hidden latency carries no regret cost.

\Result{Speculation's speedup is governed by the eval-to-overhead ratio $\rho = T_{\text{ML}}/(T_{\text{propose}}+T_{\text{judge}})$: it peaks when the ML evaluation and the agent overhead are comparable ($\rho\!\approx\!1$, the medium-model HPOBench regime), tapers off when the evaluation so dominates that the hidden overhead is a negligible fraction ($\rho\!\gg\!1$) -- where the win shifts entirely to integration, ASAP still attaining the lowest regret of all methods -- and is out of scope only when the evaluation is too cheap to hide overhead under ($\rho\!\ll\!1$).}
\textbf{Where speculation pays off.} The benefit of hiding tool$+$Judge under $T_{\text{ML}}$ is not uniform across tasks but is set by a single ratio, $\rho = T_{\text{ML}}/(T_{\text{propose}}+T_{\text{judge}})$, the cost of the ML evaluation relative to the agent overhead it can hide it under (Fig.~\ref{fig:spec-ratio}). The acceleration is largest when the two stages are comparable ($\rho\!\approx\!1$), the medium-model HPOBench regime (RandomForest, XGBoost) where hiding one of two roughly equal stages saves the most per trial. As $\rho$ grows further ($\rho\!\gg\!1$, the large-model PD1 regime where $T_{\text{ML}}$ reaches hours), the overhead is still fully hidden, but it is now a small fraction of $T_{\text{ML}}$, so the relative speedup shrinks to a few percent even though it remains free. Crucially, ASAP stays dominant even here: when speculation has little left to hide, the advantage comes \emph{from integration} rather than acceleration, and ASAP still attains the lowest regret of all methods (Fig.~\ref{fig:regret-vs-wallclock}). Only at the opposite end, $\rho\!\ll\!1$ (small data, simple models such as the sub-second trivial-eval regime of Sec.~\ref{sec:trivial-eval}), is $T_{\text{ML}}$ too short to absorb the overhead, so speculation cannot help, the out-of-scope regime. ASAP therefore targets the broad non-trivial-eval band, exactly the operating point of modern complex model-training HPO, and remains the best method across it whether or not speculation contributes.

\paragraph{Self-Tuner behavior.}
\Result{The Self-Tuner adapts the accept threshold $\tau$ online to task-specific values, and is latency-safe by construction: a rejected speculation re-runs only tool+Judge, never the ML evaluation.}
The Self-Tuner adapts $\tau$ online, with no hand-tuning, tracking the speculation error scale and re-deriving a task-specific tolerance. We show this on the HPOBench tasks where speculation is effective (Fig.~\ref{fig:self-evolver}); PD1 speculation is structurally near-flat (Sec.~\ref{sec:c3-speculation}) and omitted.
\textbf{(1) Online, principled adaptation.} The indigo curve (Fig.~\ref{fig:self-evolver}) is $\tau$ averaged over tasks. It starts at the warm-up $\tau=0.30$ and tightens to ${\sim}0.10$ as in-context history accumulates, tracking down toward the empirical prediction-error scale (green mean $|\text{rel-err}|$, ${\sim}0.05$--$0.08$). The mechanism makes this automatic: $\tau$ is set to a quantile of recent relative errors, so as predictions sharpen the tolerance contracts. No fixed $\tau$ reproduces this: a tight one (e.g.\ $0.05$) over-rejects during cold start, discarding early savings; a loose one (e.g.\ $0.30$) never tightens. The Self-Tuner threads both without task- or stage-specific tuning.
\textbf{(2) Task-dependent convergence.} This is not a fixed schedule in disguise. The $\pm$std band and right-margin density (Fig.~\ref{fig:self-evolver}) show that, from the identical warm-up $\tau=0.30$, the converged $\tau$ fans out to task-specific values spanning $[0.03,0.25]$, nearly an order of magnitude. Since the only input is each task's own realized error distribution, this spread reads out how predictable each task is: tasks whose outcomes the prediction nails settle to a tight $\tau$, noisier tasks to a loose one. No global $\tau$ fits all, and the Self-Tuner finds each task's value with no tuning.
\textbf{(3) Latency-safe by construction.} A rejected speculation re-runs only the cheap tool$+$Judge step, never $T_{\text{ML}}$ (Sec.~\ref{sec:c3-speculation}), so the worst case is exactly the serial baseline. A too-tight $\tau$ only forgoes wall-clock savings; a too-loose $\tau$ admits predictions within relative error $\tau$, which can shift the committed config, so regret is protected by keeping $\tau$ near the prediction-error scale, which the Self-Tuner does automatically. The Self-Tuner is thus tuning-free insurance: it captures the savings where prediction is reliable and tightens to protect decision quality where it is not.

\subsection{RQ3: Case Studies for Interpreting Optimization Behaviors}
\label{sec:rq3-analysis}

RQ1--RQ2 quantified \emph{that} ASAP optimizes better and uses its pool; here we interpret \emph{how}. Each proposer concentrates in its own biased sub-region of the hyperparameter space (sometimes far from the optimum, sometimes only near it), yet the discriminative Judge switches between proposers across iterations and reaches the optimum in either regime.

\paragraph{Every single proposer concentrates in a sub-region; the Judge integrates across them.}
\Result{Each single proposer concentrates in a biased sub-region, so none reliably reaches the optimum; by switching between proposers, only the Judge-integrated ASAP queries the optimum in all three case studies.}
ASAP does not \emph{generate} hyperparameters: every committed query is selected by the discriminative Judge from the pool (seven BO tools, GP, TPE, Optuna, SKOPT, TuRBO, SMAC, DNGO, plus the LLAMBO ICL proposer). Fig.~\ref{fig:case-study} shows three case studies spanning different benchmarks, models, and failure modes. Each panel projects the search onto two hyperparameters and draws, for four representative proposers (GP, TPE, TuRBO, LLAMBO), a KDE contour of their queries plus a sparse same-color scatter, and ASAP's queries are the larger green dots. The pattern is consistent: each proposer's queries cluster tightly inside its own region, and only ASAP's queries reach the corner holding the gold-star optimum.
\textbf{(a) PD1/uniref50/Transformer.} BO tools sit smaller \texttt{opt\_momentum}, well below the optimum at ${\approx}1.0$; LLAMBO reaches the highest momentum but its \texttt{lr\_decay} support stops near $0.55$, short of the optimum at $0.99$. No single proposer reaches the star; ASAP's queries span from LLAMBO's high-momentum band to the upper-right corner that holds it.
\textbf{(b) PD1/cifar100/Wide-ResNet.} The BO tools concentrate at \texttt{lr\_decay}${\approx}0.7$--$0.95$, just short of the optimum at $0.99$; LLAMBO is broader in \texttt{lr\_initial} but drifts to lower \texttt{lr\_decay}${\approx}0.2$--$0.55$. ASAP queries past the BO cluster to the star at $0.99$ on both seeds, also covering the LLAMBO band.
\textbf{(c) HPOBench/167119/XGBoost.} On a different benchmark and search space, the pattern reappears: the BO tools concentrate at moderate \texttt{eta} (${\approx}0.05$--$0.7$); LLAMBO's region grazes the star (high \texttt{max\_depth}${=}50$, low \texttt{eta}) but its queries do not converge to it. ASAP covers both regions and reaches the star on both seeds.

Across all three panels the same mechanism holds: each proposer is biased toward a sub-region, and the Judge integrates across the pool (committing one proposer's proposal one round, another's the next) so that ASAP reaches the star even when no single proposer's region does (a, b) or when one grazes it imprecisely (c). The escape is attributable to integration itself (Sec.~\ref{sec:integrate-vs-replace}).
\section{Conclusion}
\label{Sec:conclusion}
We presented \textbf{ASAP}, an agent-system co-design for hyperparameter optimization. It targets two weaknesses of prior LLM-HPO: each method bets on a \emph{single} inductively-biased proposer, fragile when a task drifts from that prior, and optimizes iteration count while per-round LLM and tool overhead dominates \emph{real} wall-clock. Module~1 integrates a diverse pool of optimizers under one LLM-as-Judge, so complementary biases jointly cover problems no single proposer handles; Module~2 re-architects the loop around end-to-end wall-clock via a prefix-stable prompt, cross-iteration speculation with safe fallback, and an off-critical-path Self-Tuner. Across 28 combinations on HPOBench and PD1, ASAP reaches lower regret per unit wall-clock than every baseline, and analysis show it stays in the top group on every landscape axis, the regime where single-bias baselines each collapse on at least one. Ablations attribute the regret floor to Module~1 (integration) and the wall-clock saving to Module~2 (speculation and Self-Tuner).

\section{Limitations}
ASAP has two main limitations. First, the speculation gain scales with the evaluation cost it hides under: when $T_{\text{ML}}$ is very cheap (below roughly 5\,s), the hiding window is too small to overlap most of the propose+Judge overhead, so speculation helps little. Second, the KV-cache-aware prompt assumes the serving engine reuses the KV cache of token-identical prefixes; where such reuse is unsupported, this benefit degrades.

\bibliographystyle{ACM-Reference-Format}
\bibliography{custom}

@article{JMLR:v13:bergstra12a,
  author  = {James Bergstra and Yoshua Bengio},
  title   = {Random Search for Hyper-Parameter Optimization},
  journal = {Journal of Machine Learning Research},
  year    = {2012},
  volume  = {13},
  number  = {10},
  pages   = {281--305},
  url     = {http://jmlr.org/papers/v13/bergstra12a.html}
}

@misc{skopt,
  author       = {scikit-optimize},
  howpublished = {\url{https://scikit-optimize.github.io/stable/}},
}

@misc{akiba2019optunanextgenerationhyperparameteroptimization,
      title={Optuna: A Next-generation Hyperparameter Optimization Framework}, 
      author={Takuya Akiba and Shotaro Sano and Toshihiko Yanase and Takeru Ohta and Masanori Koyama},
      year={2019},
      eprint={1907.10902},
      archivePrefix={arXiv},
      primaryClass={cs.LG},
      url={https://arxiv.org/abs/1907.10902}, 
}

@inproceedings{NIPS1995_7cce53cf,
 author = {Williams, Christopher and Rasmussen, Carl},
 booktitle = {Advances in Neural Information Processing Systems},
 editor = {D. Touretzky and M.C. Mozer and M. Hasselmo},
 pages = {},
 publisher = {MIT Press},
 title = {Gaussian Processes for Regression},
 url = {https://proceedings.neurips.cc/paper_files/paper/1995/file/7cce53cf90577442771720a370c3c723-Paper.pdf},
 volume = {8},
 year = {1995}
}

@misc{watanabe2026treestructuredparzenestimatorunderstanding,
      title={Tree-Structured Parzen Estimator: Understanding Its Algorithm Components and Their Roles for Better Empirical Performance}, 
      author={Shuhei Watanabe},
      year={2026},
      eprint={2304.11127},
      archivePrefix={arXiv},
      primaryClass={cs.LG},
      url={https://arxiv.org/abs/2304.11127}, 
}

@inproceedings{NIPS2014_b610047c,
 author = {Ding, Nan and Fang, Youhan and Babbush, Ryan and Chen, Changyou and Skeel, Robert D. and Neven, Hartmut},
 booktitle = {Advances in Neural Information Processing Systems},
 editor = {Z. Ghahramani and M. Welling and C. Cortes and N. Lawrence and K. Weinberger},
 pages = {},
 publisher = {Curran Associates, Inc.},
 title = {Bayesian Sampling Using Stochastic Gradient Thermostats},
 url = {https://proceedings.neurips.cc/paper_files/paper/2014/file/b610047c85e73cb7ec04fd36ec503f93-Paper.pdf},
 volume = {27},
 year = {2014}
}

@misc{liu2024largelanguagemodelsenhance,
      title={Large Language Models to Enhance Bayesian Optimization}, 
      author={Tennison Liu and Nicolás Astorga and Nabeel Seedat and Mihaela van der Schaar},
      year={2024},
      eprint={2402.03921},
      archivePrefix={arXiv},
      primaryClass={cs.LG},
      url={https://arxiv.org/abs/2402.03921}, 
}

@misc{liu2025largelanguagemodelagent,
      title={Large Language Model Agent for Hyper-Parameter Optimization}, 
      author={Siyi Liu and Chen Gao and Yong Li},
      year={2025},
      eprint={2402.01881},
      archivePrefix={arXiv},
      primaryClass={cs.LG},
      url={https://arxiv.org/abs/2402.01881}, 
}

@misc{li2024studybayesianneuralnetwork,
      title={A Study of Bayesian Neural Network Surrogates for Bayesian Optimization}, 
      author={Yucen Lily Li and Tim G. J. Rudner and Andrew Gordon Wilson},
      year={2024},
      eprint={2305.20028},
      archivePrefix={arXiv},
      primaryClass={cs.LG},
      url={https://arxiv.org/abs/2305.20028}, 
}

@misc{tornede2024automlagelargelanguage,
      title={AutoML in the Age of Large Language Models: Current Challenges, Future Opportunities and Risks}, 
      author={Alexander Tornede and Difan Deng and Theresa Eimer and Joseph Giovanelli and Aditya Mohan and Tim Ruhkopf and Sarah Segel and Daphne Theodorakopoulos and Tanja Tornede and Henning Wachsmuth and Marius Lindauer},
      year={2024},
      eprint={2306.08107},
      archivePrefix={arXiv},
      primaryClass={cs.LG},
      url={https://arxiv.org/abs/2306.08107}, 
}

@misc{hong2024datainterpreterllmagent,
      title={Data Interpreter: An LLM Agent For Data Science}, 
      author={Sirui Hong and Yizhang Lin and Bang Liu and Bangbang Liu and Binhao Wu and Ceyao Zhang and Chenxing Wei and Danyang Li and Jiaqi Chen and Jiayi Zhang and Jinlin Wang and Li Zhang and Lingyao Zhang and Min Yang and Mingchen Zhuge and Taicheng Guo and Tuo Zhou and Wei Tao and Xiangru Tang and Xiangtao Lu and Xiawu Zheng and Xinbing Liang and Yaying Fei and Yuheng Cheng and Zhibin Gou and Zongze Xu and Chenglin Wu},
      year={2024},
      eprint={2402.18679},
      archivePrefix={arXiv},
      primaryClass={cs.AI},
      url={https://arxiv.org/abs/2402.18679}, 
}

@misc{guo2024dsagentautomateddatascience,
      title={DS-Agent: Automated Data Science by Empowering Large Language Models with Case-Based Reasoning}, 
      author={Siyuan Guo and Cheng Deng and Ying Wen and Hechang Chen and Yi Chang and Jun Wang},
      year={2024},
      eprint={2402.17453},
      archivePrefix={arXiv},
      primaryClass={cs.LG},
      url={https://arxiv.org/abs/2402.17453}, 
}

@article{guo2025mtsql,
  title={MTSQL-R1: Towards Long-Horizon Multi-Turn Text-to-SQL via Agentic Training},
  author={Guo, Taicheng and Wang, Hai and Liu, ChaoChun and Golalikhani, Mohsen and Chen, Xin and Zhang, Xiangliang and Reddy, Chandan K},
  journal={arXiv preprint arXiv:2510.12831},
  year={2025}
}

@article{guo2026autollmresearch,
  title={AutoLLMResearch: Training Research Agents for Automating LLM Experiment Configuration--Learning from Cheap, Optimizing Expensive},
  author={Guo, Taicheng and Chawla, Nitesh V and Wiest, Olaf and Zhang, Xiangliang},
  journal={arXiv preprint arXiv:2605.11518},
  year={2026}
}

@misc{chi2024selatreesearchenhancedllm,
      title={SELA: Tree-Search Enhanced LLM Agents for Automated Machine Learning}, 
      author={Yizhou Chi and Yizhang Lin and Sirui Hong and Duyi Pan and Yaying Fei and Guanghao Mei and Bangbang Liu and Tianqi Pang and Jacky Kwok and Ceyao Zhang and Bang Liu and Chenglin Wu},
      year={2024},
      eprint={2410.17238},
      archivePrefix={arXiv},
      primaryClass={cs.AI},
      url={https://arxiv.org/abs/2410.17238}, 
}

@misc{zhang2024usinglargelanguagemodels,
      title={Using Large Language Models for Hyperparameter Optimization}, 
      author={Michael R. Zhang and Nishkrit Desai and Juhan Bae and Jonathan Lorraine and Jimmy Ba},
      year={2024},
      eprint={2312.04528},
      archivePrefix={arXiv},
      primaryClass={cs.LG},
      url={https://arxiv.org/abs/2312.04528}, 
}

@misc{leviathan2023fastinferencetransformersspeculative,
      title={Fast Inference from Transformers via Speculative Decoding}, 
      author={Yaniv Leviathan and Matan Kalman and Yossi Matias},
      year={2023},
      eprint={2211.17192},
      archivePrefix={arXiv},
      primaryClass={cs.LG},
      url={https://arxiv.org/abs/2211.17192}, 
}

@misc{ye2026speculativeactionslosslessframework,
      title={Speculative Actions: A Lossless Framework for Faster Agentic Systems}, 
      author={Naimeng Ye and Arnav Ahuja and Georgios Liargkovas and Yunan Lu and Kostis Kaffes and Tianyi Peng},
      year={2026},
      eprint={2510.04371},
      archivePrefix={arXiv},
      primaryClass={cs.AI},
      url={https://arxiv.org/abs/2510.04371}, 
}

@misc{guan2025dynamicspeculativeagentplanning,
      title={Dynamic Speculative Agent Planning}, 
      author={Yilin Guan and Qingfeng Lan and Sun Fei and Dujian Ding and Devang Acharya and Chi Wang and William Yang Wang and Wenyue Hua},
      year={2025},
      eprint={2509.01920},
      archivePrefix={arXiv},
      primaryClass={cs.AI},
      url={https://arxiv.org/abs/2509.01920}, 
}

@misc{ro2025sherlockreliableefficientagentic,
      title={Sherlock: Reliable and Efficient Agentic Workflow Execution}, 
      author={Yeonju Ro and Haoran Qiu and Íñigo Goiri and Rodrigo Fonseca and Ricardo Bianchini and Aditya Akella and Zhangyang Wang and Mattan Erez and Esha Choukse},
      year={2025},
      eprint={2511.00330},
      archivePrefix={arXiv},
      primaryClass={cs.MA},
      url={https://arxiv.org/abs/2511.00330}, 
}

@misc{wu2026evolverselfevolvingllmagents,
      title={EvolveR: Self-Evolving LLM Agents through an Experience-Driven Lifecycle}, 
      author={Rong Wu and Xiaoman Wang and Jianbiao Mei and Pinlong Cai and Daocheng Fu and Cheng Yang and Licheng Wen and Xuemeng Yang and Yufan Shen and Yuxin Wang and Botian Shi},
      year={2026},
      eprint={2510.16079},
      archivePrefix={arXiv},
      primaryClass={cs.CL},
      url={https://arxiv.org/abs/2510.16079}, 
}

@misc{chen2025multiagentevolvellmselfimprove,
      title={Multi-Agent Evolve: LLM Self-Improve through Co-evolution}, 
      author={Yixing Chen and Yiding Wang and Siqi Zhu and Haofei Yu and Tao Feng and Muhan Zhang and Mostofa Patwary and Jiaxuan You},
      year={2025},
      eprint={2510.23595},
      archivePrefix={arXiv},
      primaryClass={cs.AI},
      url={https://arxiv.org/abs/2510.23595}, 
}

@misc{gao2026surveyselfevolvingagentswhat,
      title={A Survey of Self-Evolving Agents: What, When, How, and Where to Evolve on the Path to Artificial Super Intelligence}, 
      author={Huan-ang Gao and Jiayi Geng and Wenyue Hua and Mengkang Hu and Xinzhe Juan and Hongzhang Liu and Shilong Liu and Jiahao Qiu and Xuan Qi and Yiran Wu and Hongru Wang and Han Xiao and Yuhang Zhou and Shaokun Zhang and Jiayi Zhang and Jinyu Xiang and Yixiong Fang and Qiwen Zhao and Dongrui Liu and Qihan Ren and Cheng Qian and Zhenhailong Wang and Minda Hu and Huazheng Wang and Qingyun Wu and Heng Ji and Mengdi Wang},
      year={2026},
      eprint={2507.21046},
      archivePrefix={arXiv},
      primaryClass={cs.AI},
      url={https://arxiv.org/abs/2507.21046}, 
}

@misc{snoek2012practicalbayesianoptimizationmachine,
      title={Practical Bayesian Optimization of Machine Learning Algorithms}, 
      author={Jasper Snoek and Hugo Larochelle and Ryan P. Adams},
      year={2012},
      eprint={1206.2944},
      archivePrefix={arXiv},
      primaryClass={stat.ML},
      url={https://arxiv.org/abs/1206.2944}, 
}

@inproceedings{NIPS2011_86e8f7ab,
 author = {Bergstra, James and Bardenet, R\'{e}mi and Bengio, Yoshua and K\'{e}gl, Bal\'{a}zs},
 booktitle = {Advances in Neural Information Processing Systems},
 editor = {J. Shawe-Taylor and R. Zemel and P. Bartlett and F. Pereira and K. Weinberger},
 pages = {},
 publisher = {Curran Associates, Inc.},
 title = {Algorithms for Hyper-Parameter Optimization},
 url = {https://proceedings.neurips.cc/paper_files/paper/2011/file/86e8f7ab32cfd12577bc2619bc635690-Paper.pdf},
 volume = {24},
 year = {2011}
}

@misc{lindauer2022smac3versatilebayesianoptimization,
      title={SMAC3: A Versatile Bayesian Optimization Package for Hyperparameter Optimization}, 
      author={Marius Lindauer and Katharina Eggensperger and Matthias Feurer and André Biedenkapp and Difan Deng and Carolin Benjamins and Tim Ruhopf and René Sass and Frank Hutter},
      year={2022},
      eprint={2109.09831},
      archivePrefix={arXiv},
      primaryClass={cs.LG},
      url={https://arxiv.org/abs/2109.09831}, 
}

@misc{eriksson2020scalableglobaloptimizationlocal,
      title={Scalable Global Optimization via Local Bayesian Optimization}, 
      author={David Eriksson and Michael Pearce and Jacob R Gardner and Ryan Turner and Matthias Poloczek},
      year={2020},
      eprint={1910.01739},
      archivePrefix={arXiv},
      primaryClass={cs.LG},
      url={https://arxiv.org/abs/1910.01739}, 
}

@misc{snoek2015scalablebayesianoptimizationusing,
      title={Scalable Bayesian Optimization Using Deep Neural Networks}, 
      author={Jasper Snoek and Oren Rippel and Kevin Swersky and Ryan Kiros and Nadathur Satish and Narayanan Sundaram and Md. Mostofa Ali Patwary and Prabhat and Ryan P. Adams},
      year={2015},
      eprint={1502.05700},
      archivePrefix={arXiv},
      primaryClass={stat.ML},
      url={https://arxiv.org/abs/1502.05700}, 
}

@misc{eggensperger2022hpobenchcollectionreproduciblemultifidelity,
      title={HPOBench: A Collection of Reproducible Multi-Fidelity Benchmark Problems for HPO}, 
      author={Katharina Eggensperger and Philipp Müller and Neeratyoy Mallik and Matthias Feurer and René Sass and Aaron Klein and Noor Awad and Marius Lindauer and Frank Hutter},
      year={2022},
      eprint={2109.06716},
      archivePrefix={arXiv},
      primaryClass={cs.LG},
      url={https://arxiv.org/abs/2109.06716}, 
}

@misc{wang2024pretrainedgaussianprocessesbayesian,
      title={Pre-trained Gaussian Processes for Bayesian Optimization}, 
      author={Zi Wang and George E. Dahl and Kevin Swersky and Chansoo Lee and Zachary Nado and Justin Gilmer and Jasper Snoek and Zoubin Ghahramani},
      year={2024},
      eprint={2109.08215},
      archivePrefix={arXiv},
      primaryClass={cs.LG},
      url={https://arxiv.org/abs/2109.08215}, 
}

@misc{kerschke2017comprehensivefeaturebasedlandscapeanalysis,
      title={Comprehensive Feature-Based Landscape Analysis of Continuous and Constrained Optimization Problems Using the R-Package flacco}, 
      author={Pascal Kerschke},
      year={2017},
      eprint={1708.05258},
      archivePrefix={arXiv},
      primaryClass={stat.ML},
      url={https://arxiv.org/abs/1708.05258}, 
}

@article{wolpert2002no,
  title={No free lunch theorems for optimization},
  author={Wolpert, David H and Macready, William G},
  journal={IEEE transactions on evolutionary computation},
  volume={1},
  number={1},
  pages={67--82},
  year={2002},
  publisher={IEEE}
}

\appendix
\clearpage
\section*{Appendix}
\label{sec:Appendix}

\section{Implementation Details}
\label{sec:impl-details}
This section collects everything needed to reproduce our experiments: the benchmark tasks we evaluate on (Sec.~\ref{sec:bench-details}), the full hyperparameter search spaces (Sec.~\ref{sec:hyper_space}), and the prompt design used by the LLM-as-Judge inside Module~1 (Sec.~\ref{sec:problem-context}).

\subsection{Benchmark Details}
\label{sec:bench-details}

Table~\ref{tab:bench-tasks} lists the 28 (task, model) combinations used in our evaluation. ``ML Eval Time'' is the median measured wall-clock cost of a single configuration evaluation: real training time for HPOBench, and pre-computed GPU training wall (the cost a practitioner actually pays per trial in deep-learning HPO) for PD1, which the benchmark surfaces from $\sim$50k recorded training runs. ``Objective'' is the metric the benchmark returns (validation accuracy for HPOBench classification; validation error rate from the pre-computed training run for PD1). All tasks are pure HPO with a 4-D search space; full ranges appear in Sec.~\ref{sec:hyper_space}.

The pool deliberately covers (i) classical tabular learners under real training cost (HPOBench / XGBoost, RandomForest), (ii) modern CNN architectures of practical interest (Wide-ResNet, ResNet-50, Simple-CNN, Max-Pooling-CNN) on standard vision benchmarks at the batch sizes used by their authors, and (iii) modern Transformer-style training on the kind of workloads that dominate today's HPO cost: language modeling (LM1B), machine translation (WMT En--De), and protein modeling (Uniref50). Per-trial cost spans four orders of magnitude, from $\sim$9\,s for the cheapest HPOBench task to $\sim$48\,h for the most expensive (WMT En--De translation) and $\sim$25\,h for ImageNet / ResNet-50, so every operating regime relevant to ASAP's walltime-centered design appears in the evaluation.

\begin{table*}[h]
\centering
\footnotesize
\setlength{\tabcolsep}{4pt}
\begin{tabular}{llllrl}
\hline
\textbf{Benchmark} & \textbf{Domain} & \textbf{Dataset} & \textbf{Model} & \textbf{ML Eval Time (s)} & \textbf{Objective} \\ \hline
\multicolumn{6}{l}{\textit{HPOBench (raw mode: real training of sklearn-style tabular models on OpenML)}} \\
HPOBench & Tabular & jasmine (OpenML 167119, 27k$\times$6, 3-class)    & RandomForest & 9.5   & Val.\ accuracy \\
HPOBench & Tabular & jasmine                                             & XGBoost      & 37    & Val.\ accuracy \\
HPOBench & Tabular & adult (OpenML 7592, 29k$\times$105, binary)         & RandomForest & 13    & Val.\ accuracy \\
HPOBench & Tabular & adult                                                & XGBoost      & 19    & Val.\ accuracy \\
HPOBench & Tabular & high-dim (OpenML 168908, 3.3k$\times$1646)          & RandomForest & 29    & Val.\ accuracy \\
HPOBench & Tabular & high-dim                                             & XGBoost      & 238   & Val.\ accuracy \\
HPOBench & Tabular & madeline (OpenML 167120, 58k$\times$21, binary)     & XGBoost      & 30    & Val.\ accuracy \\
HPOBench & Tabular & madeline                                             & RandomForest & 85    & Val.\ accuracy \\
HPOBench & Tabular & volkert (OpenML 168329, 39k$\times$27, 100-class)   & RandomForest & 148   & Val.\ accuracy \\ \hline
\multicolumn{6}{l}{\textit{PD1 (replay of $\sim$50k pre-computed deep-learning HPO runs; ``Eval Time'' = pre-computed GPU training wall per trial)}} \\
PD1 & Vision   & Fashion-MNIST (60k$\times$28$^2$, 10-class)               & Max-Pooling-CNN (bs=2048)  & 116      & Val.\ error \\
PD1 & Vision   & MNIST (60k$\times$28$^2$, 10-class)                       & Max-Pooling-CNN (bs=2048)  & 120      & Val.\ error \\
PD1 & Vision   & Fashion-MNIST                                              & Simple-CNN (bs=2048)       & 142      & Val.\ error \\
PD1 & Vision   & MNIST                                                      & Simple-CNN (bs=2048)       & 166      & Val.\ error \\
PD1 & Vision   & Fashion-MNIST                                              & Max-Pooling-CNN (bs=256)   & 519      & Val.\ error \\
PD1 & Vision   & MNIST                                                      & Max-Pooling-CNN (bs=256)   & 547      & Val.\ error \\
PD1 & Vision   & Fashion-MNIST                                              & Simple-CNN (bs=256)        & 592      & Val.\ error \\
PD1 & Vision   & MNIST                                                      & Simple-CNN (bs=256)        & 718      & Val.\ error \\
PD1 & Vision   & CIFAR-100 (50k$\times$32$^2$, 100-class)                  & Wide-ResNet (bs=2048)      & 1{,}560  & Val.\ error \\
PD1 & Vision   & CIFAR-10  (50k$\times$32$^2$, 10-class)                   & Wide-ResNet (bs=2048)      & 2{,}338  & Val.\ error \\
PD1 & Vision   & CIFAR-100                                                  & Wide-ResNet (bs=256)       & 2{,}484  & Val.\ error \\
PD1 & Vision   & SVHN (73k$\times$32$^2$, 10-class)                        & Wide-ResNet (bs=1024)      & 2{,}724  & Val.\ error \\
PD1 & Vision   & CIFAR-10                                                   & Wide-ResNet (bs=256)       & 3{,}934  & Val.\ error \\
PD1 & Vision   & SVHN                                                       & Wide-ResNet (bs=256)       & 4{,}157  & Val.\ error \\
PD1 & Vision   & ImageNet (1.3M$\times$224$^2$, 1000-class)                & ResNet-50 (bs=512)         & 78{,}900 & Val.\ error \\
PD1 & Protein  & Uniref50 (protein sequences)                              & Transformer (bs=128)       & 84{,}417 & Val.\ error \\
PD1 & Vision   & ImageNet                                                   & ResNet-50 (bs=256)         & 90{,}517 & Val.\ error \\
PD1 & Language & LM1B (billion-word language modeling)                     & Transformer (bs=2048)      & 110{,}293& Val.\ error \\
PD1 & Language & WMT En--De (machine translation)                          & xFormer-Translate (bs=64)  & 172{,}699& Val.\ error \\ \hline
\end{tabular}
\caption{The 28 (task, model) combinations used in our evaluation. ``ML Eval Time'' is the median measured wall-clock cost of a single configuration evaluation: real model training for HPOBench, and the pre-computed GPU training wall replayed by PD1 (the actual cost a practitioner pays per trial in deep-learning HPO). All tasks are pure HPO with a 4-D search space (full ranges in Sec.~\ref{sec:hyper_space}). The pool spans classical tabular learners (RandomForest, XGBoost), modern CNNs (Wide-ResNet, ResNet-50, Simple/Max-Pooling-CNN), and modern Transformers (language modeling, machine translation, protein), with per-trial cost ranging over four orders of magnitude ($\sim$9\,s to $\sim$48\,h), entirely within the non-trivial-eval regime that ASAP's walltime-centered design targets.}
\label{tab:bench-tasks}
\end{table*}

\subsection{Hyperparameter Search Spaces}
\label{sec:hyper_space}

\paragraph{HPOBench (XGBoost / RandomForest on OpenML tabular tasks).}
All HPOBench combos in Table~\ref{tab:bench-tasks} share a 4-D search space per model:
\begin{itemize}[nolistsep, leftmargin=*]
    \item \textbf{XGBoost}: \{colsample\_bytree: [linear, 0.1, 1.0], eta: [log, $2^{-10}$, 1.0], max\_depth: [log, 1, 50], reg\_lambda: [log, $2^{-10}$, 1024]\}
    \item \textbf{RandomForest}: \{max\_depth: [log, 1, 50], max\_features: [linear, 0.0, 1.0], min\_samples\_leaf: [linear, 1, 20], min\_samples\_split: [log, 2, 128]\}
\end{itemize}

\paragraph{PD1 (deep-learning HPO replay).}
All PD1 architectures (Wide-ResNet, ResNet-50, Simple-CNN, Max-Pooling-CNN, Transformer / xFormer-Translate) share the same 4-D optimizer search space exposed by the PD1 study schema:
\begin{itemize}[nolistsep, leftmargin=*]
    \item \{lr\_initial\_value: [log, $10^{-5}$, 10.0], lr\_power: [linear, 0.1, 2.0], lr\_decay\_steps\_factor: [linear, 0.01, 1.0], opt\_momentum: [logit, $10^{-4}$, 0.9999]\}
\end{itemize}
PD1 trains the model with these optimizer hyperparameters; the dataset and architecture identify the study, and each lookup returns the recorded best validation error of the closest precomputed trial in HP space.

\subsection{LLM-as-Judge Prompt Design}
\label{sec:problem-context}
The ICL Judge in Sec.~\ref{sec:integrated-loop} conditions on a textual problem context built from the \emph{meta-features} of the target HPO task. The context includes the ML model name and task type (classification or regression) together with key dataset attributes: number of samples, number of features, feature types, inter-feature correlations, and feature-label correlations. Dataset names are deliberately omitted to mitigate memorization risk in the LLM's pretraining corpus. On top of this context, the Judge prompt frames the prediction task in an \emph{inductive} few-shot regression style: given a list of previously evaluated (HP configuration, performance) pairs and a candidate configuration, the model predicts the candidate's performance, which the outer loop turns into an Expected Improvement score from $S$ sampled predictions per candidate (Sec.~\ref{sec:integrated-loop}). The complete template, including system message, few-shot block, candidate list, and response format, is included with the released code.

\section{Additional Experimental Results}
\label{sec:add-results}
This section reports auxiliary experiments that complement the main results in Sec.~\ref{Sec:exp}: ASAP's behavior on the \emph{trivial-eval} regime that the main scope excludes (Sec.~\ref{sec:trivial-eval}), and a breakdown of regret across qualitatively different optimization landscapes (Sec.~\ref{sec:landscapes}).

\subsection{Performance on Trivial-Eval ML Experiments ($<5$\,s)}
\label{sec:trivial-eval}

The main paper restricts to HPO tasks where per-evaluation cost is non-trivial ($\geq 5$\,s); under that regime ASAP's walltime advantage compounds because speculation hides per-round LLM/tool overhead under the ML evaluation. For completeness, we report results on the \emph{trivial-eval} regime where each ML evaluation is sub-second (classical sklearn HPO on small tabular data, Bayesmark). Here speculation cannot meaningfully hide the agent's overhead (per-round wall is dominated by the LLM/tool stack rather than the model fit), so the design assumption that motivates ASAP no longer holds, and we expect cheap model-free BO baselines (e.g., TPE) to be highly competitive on the walltime axis. This experiment quantifies that boundary and grounds the limitations discussion.

\begin{figure*}[t]
\centering
\begin{subfigure}[t]{0.48\linewidth}
    \centering
    \includegraphics[width=\linewidth]{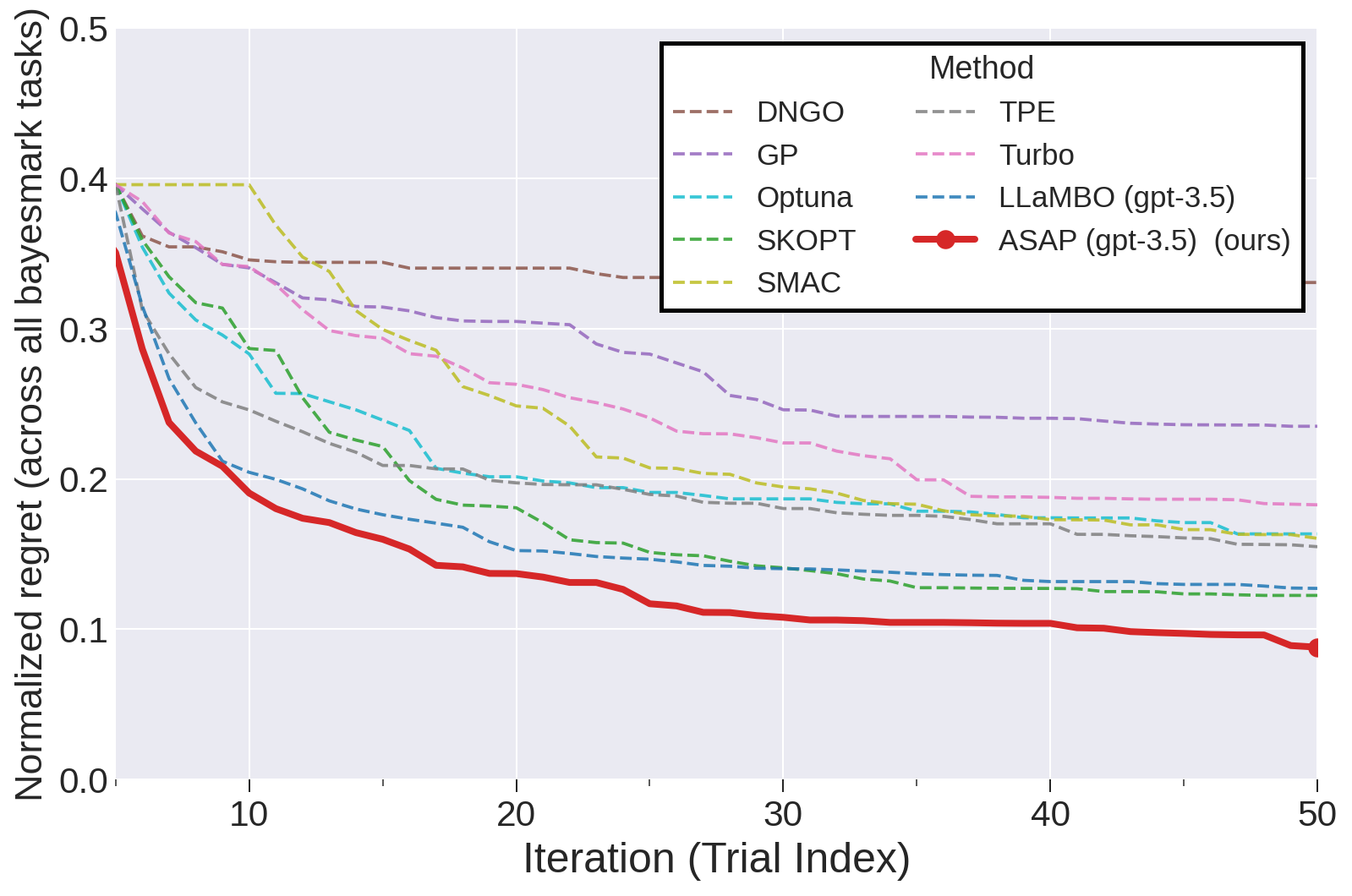}
    \caption{Iteration axis (50 trials, 3 seeds). ASAP leads throughout.}
    \label{fig:trivial-iter}
\end{subfigure}\hfill
\begin{subfigure}[t]{0.48\linewidth}
    \centering
    \includegraphics[width=\linewidth]{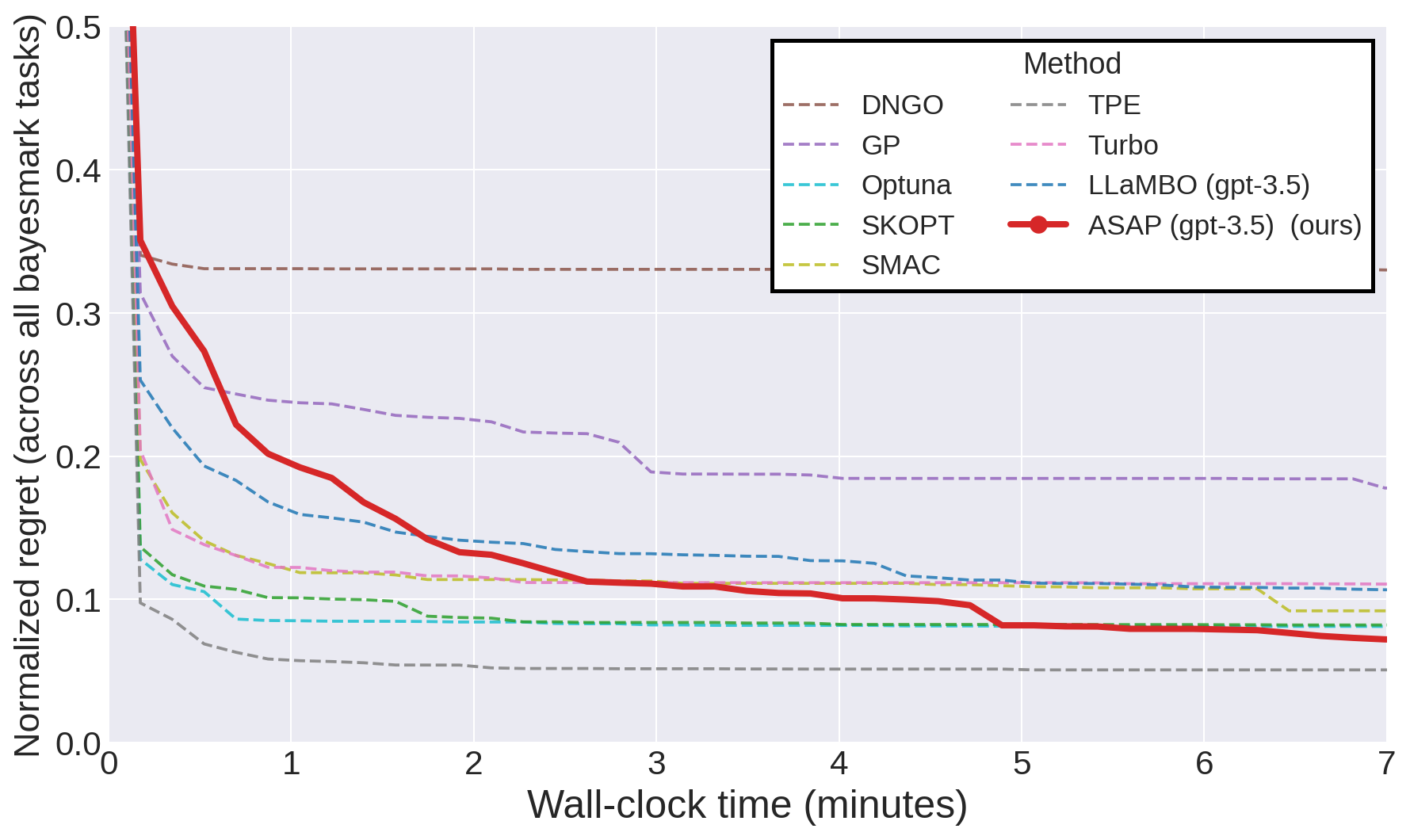}
    \caption{Walltime axis on a fair 7\,min budget (BO: 5 init + 500 trials, 3 seeds; LLM: 5 init + 50 trials, 3 seeds). TPE wins ($\approx 0.05$); ASAP ($\approx 0.075$, red) still beats LLAMBO and most BOs.}
    \label{fig:trivial-wall}
\end{subfigure}
\caption{Trivial-eval (sub-second sklearn HPO) on the full Bayesmark grid. ASAP (red, ours) retains its per-iteration advantage but loses the walltime axis to model-free BOs that fit hundreds of trials inside ASAP's 7\,min LLM-bound window. This is the boundary at which the speculation-driven walltime story breaks down, exactly outside the regime the main paper targets.}
\label{fig:trivial-eval}
\end{figure*}

\paragraph{Setup.} We use the full original Bayesmark grid: 4 CLF datasets (\textit{breast}, \textit{digits}, \textit{iris}, \textit{wine}) and 4 REG datasets (\textit{diabetes}, \textit{Griewank}, \textit{KTablet}, \textit{Rosenbrock}), each paired with 5 sklearn models (\textit{ada}, \textit{DT}, \textit{MLP-sgd}, \textit{RF}, \textit{SVM}), giving $8 \times 5 = 40$ (task, model) combos. Regret is normalized per task by the cross-method best/worst window observed across all baselines and averaged over tasks.
For the iteration plot (Fig.~\ref{fig:trivial-iter}), all methods run the same 5\,init + 50\,trial budget, and we average three seeds per method, plotting the first 50 trials. For the walltime plot (Fig.~\ref{fig:trivial-wall}), to give the BO baselines a fair wall-time budget comparable to the LLM methods' $\sim$7\,min, we re-ran every BO tool with a much larger $5\,\text{init} + 500\,\text{trials}$ budget; the LLM methods keep their original 80-trial runs since they already consume the full 7\,min window.

\paragraph{Hyperparameter Search Spaces}
The Bayesmark search space used in classical sklearn HPO:
\begin{itemize}[nolistsep, leftmargin=*]
    \item \textbf{SVM}: \{C: [log, 1, $10^3$], $\gamma$: [log, $10^{-4}$, $10^{-3}$], tolerance: [log, $10^{-5}$, $10^{-1}$] \}
    \item \textbf{DecisionTree}: \{max depth: [linear, 1, 15], min samples split: [logit, 0.01, 0.99], min samples leaf: [logit, 0.01, 0.49], min weight fraction leaf: [logit, 0.01, 0.49], max features: [logit, 0.01, 0.99], min impurity decrease: [linear, 0.0, 0.5] \}
    \item \textbf{RandomForest}: \{max depth: [linear, 1, 15], min samples split: [logit, 0.01, 0.99], min samples leaf: [logit, 0.01, 0.49], min weight fraction leaf: [logit, 0.01, 0.49], max features: [logit, 0.01, 0.99], min impurity decrease: [linear, 0.0, 0.5] \}
    \item \textbf{MLP\_SGD}: \{hidden layer sizes: [linear, 50, 200], alpha: [log, $10^{-5}$, $10^1$], batch size: [linear, 10, 250], learning rate init: [log, $10^{-5}$, $10^{-1}$], power t: [logit, 0.1, 0.9], momentum: [logit, 0.001, 0.999] \}
    \item \textbf{AdaBoost}: \{n estimators: [linear, 10, 100], learning rate: [log, $10^{-4}$, $10^1$] \}
\end{itemize}

\paragraph{Findings.}
On the iteration axis (Fig.~\ref{fig:trivial-iter}), ASAP retains its decision-quality advantage: at trial 50 ASAP's normalized regret is $\approx 0.09$, ahead of LLAMBO ($\approx 0.13$) and the best BO (SKOPT $\approx 0.12$); the gap to all other BOs is even larger. ASAP's pool-plus-Judge selection remains the most sample-efficient on a per-iteration basis.

On the walltime axis with a fair 7\,min budget (Fig.~\ref{fig:trivial-wall}), the picture changes. Trivial-eval lets model-free BOs blast through hundreds of trials per minute, and the strongest of them (TPE) actually \emph{beats} ASAP: TPE plateaus at $\approx 0.05$ regret within $\sim$1\,min and stays there, while ASAP needs $\sim$5\,min before crossing under the BO plateau and ends at $\approx 0.075$. Optuna and SKOPT ($\approx 0.085$) lead ASAP for most of the 7\,min window, though ASAP edges under them by the end. ASAP remains ahead of LLAMBO ($\approx 0.11$), TuRBO ($\approx 0.11$), SMAC ($\approx 0.09$), GP ($\approx 0.18$), and DNGO ($\approx 0.33$). This is exactly the boundary result we expect: when per-trial wall is dominated by the LLM stack and the ML fit is essentially free, a model-free BO that can issue $5\times$ more trials in the same wall window wins. The ranking inversion confirms that ASAP's design assumption, a non-trivial $T_{\mathrm{ML}}$ to hide overhead under, is load-bearing. In the main paper's non-trivial-eval regime ($T_{\mathrm{ML}} \geq 5$\,s) the same TPE that wins here is among the worst baselines, illustrating that ``best HPO algorithm'' is regime-dependent and that the trivial-eval regime is not the one practitioners actually pay for in real model training.

\subsection{Optimization-Landscape Features and Per-Task Classification}
\label{sec:landscapes}

This section documents the flacco-style~\cite{kerschke2017comprehensivefeaturebasedlandscapeanalysis} landscape features that drive Fig.~\ref{fig:landscape-radar} in the main paper. We compute every feature \emph{from existing logs} (no extra evaluations) by pooling the (config, value) pairs that every method already produced on each task. Concretely, for a single (task, model) combo $T$ we collect all $35 \times 3 \times 9 = 945$ evaluated points across the 7 BO baselines, LLAMBO, and ASAP (Ours) (3 seeds each), normalize each configuration $x \in \mathbb{R}^4$ to $[0,1]^4$ using the search-space bounds in Sec.~\ref{sec:hyper_space} (with the correct linear / log / logit warping per hyperparameter), and treat the resulting $(X, y)$ as a sample of the optimization landscape on which we evaluate the seven scalar features below. Aliases: $N_{\text{pts}} = |y|$, $\bar x_i$ standardized $x_i$.

\paragraph{Smooth or Rugged.}
\textbf{(1) Dispersion ratio} \texttt{disp\_ratio\_05} is
\[
  \texttt{disp\_ratio\_05} = \mathrm{disp}(X_{\text{top 5\%}}) \,/\, \mathrm{disp}(X),
\]
where $\mathrm{disp}(\cdot)$ is the mean pairwise Euclidean distance among the listed points and $X_{\text{top 5\%}}$ are the top 5\% by $y$. Smaller values indicate the high-fitness region is concentrated (smooth), larger values indicate the top points are spread over multiple basins (rugged).
\textbf{(2) Information content} \texttt{ic\_hmax} is the Shannon entropy of bigrams of $\mathrm{sign}(\Delta y_t)$ along a greedy nearest-neighbor walk through $X$, normalized to $[0,1]$. Larger $\Rightarrow$ value sign flips more often along the walk $\Rightarrow$ rugged.

\paragraph{Multi-Modal.}
\textbf{(3) Number of peaks} \texttt{n\_peaks} is the number of local maxima of a smoothed kernel-density estimate of the value histogram (proximity-pruned with prominence $\geq 0.05\,\hat f_{\max}$). More peaks $\Rightarrow$ more modes.
\textbf{(4) Kurtosis} of $y$ (Pearson, not excess). High kurtosis indicates that the value distribution has heavy outlier modes alongside a dominant basin.

\paragraph{Deceptive.}
\textbf{(5) Fitness-distance correlation} \texttt{fdc} is
\[
  \texttt{fdc} = |\rho_{\mathrm{Spearman}}(d_i, -y_i)|,
\]
where $d_i = \lVert x_i - x^\ast \rVert_2$ and $x^\ast = \arg\max y$. Smaller values mean the value gives no signal about distance to the best observed point (deceptive); larger values mean the response monotonically guides search toward the optimum.

\paragraph{Anisotropic.}
\textbf{(6)} \texttt{coef\_max\_by\_min} is
\[
  \texttt{coef\_max\_by\_min} = \max_j |\beta_j| \,/\, \min_j |\beta_j|
\]
from an OLS fit of $\bar y$ on $\bar X$. Larger $\Rightarrow$ the response depends much more strongly on some hyperparameters than others (anisotropic sensitivity).
\textbf{(7)} \texttt{quad\_cond} is the condition number of the $4 \times 4$ Hessian $H$ of the simple quadratic fit
\[
  \bar y = a + b^\top \bar x + \tfrac{1}{2}\, \bar x^\top H \bar x
\]
(Hessian symmetrized; absolute eigenvalues; condition number $|\lambda|_{\max}/|\lambda|_{\min}$). Larger $\Rightarrow$ axis-aligned curvatures differ more strongly $\Rightarrow$ basin is more elongated.

\paragraph{Classification into the four landscape axes.}
For each feature we $z$-score across the 28 tasks (after a $\log_{10}$ transform for the two heavy-tailed anisotropy features). The four landscape axes used in Fig.~\ref{fig:landscape-radar} are then the simple combinations
\begin{equation}
\begin{aligned}
\mathrm{Rugged}      &= \tfrac{1}{2}(z_{\text{ic\_hmax}} + z_{\text{disp\_ratio\_05}}), \\
\mathrm{Multi\text{-}Modal} &= \tfrac{1}{2}(z_{\text{n\_peaks}} + z_{\text{kurtosis}}), \\
\mathrm{Deceptive}   &= -z_{\text{fdc}}, \\
\mathrm{Anisotropic} &= \tfrac{1}{2}(z_{\text{coef\_max\_by\_min}} + z_{\text{quad\_cond}}).
\end{aligned}
\label{eq:landscape-axes}
\end{equation}
For each axis we keep the 14 tasks (top 50\%) whose axis score is largest (the ``strongly expressing'' subset), and report each method's mean normalized regret at iteration 35 averaged over those tasks (and 3 seeds each). In Fig.~\ref{fig:landscape-radar} the radial axis is inverted so that $0$ regret sits at the outer ring and larger regret pulls lines toward the centre (``outer $=$ better''). Membership is non-exclusive: tasks can express multiple axes.

\paragraph{Per-task feature values and membership.}
Table~\ref{tab:landscape-features} lists the seven feature values for every task in our paper scope and marks which of the four landscape axes that task belongs to. The right-hand four columns make the classification reproducible end-to-end from this table alone.

\begin{table*}[h]
\centering
\scriptsize
\setlength{\tabcolsep}{4pt}
\begin{tabular}{lrrrrrrrcccc}
\hline
\textbf{Task / Model} & \textbf{disp\_05} & \textbf{ic\_hmax} & \textbf{n\_pk} & \textbf{kurt} & \textbf{fdc} & \textbf{coef$_{\max/\min}$} & \textbf{quad.cond} & \textbf{R} & \textbf{M} & \textbf{D} & \textbf{A} \\ \hline
adult/RF & 0.71 & 0.96 & 1 & 7.37 & 0.57 & 34.0 & 7.80 & \checkmark & \checkmark &  &  \\
adult/xgboost & 1.03 & 0.95 & 1 & 20.64 & 0.07 & 11.0 & 33.7 & \checkmark & \checkmark & \checkmark & \checkmark \\
high-dim/RF & 0.51 & 0.88 & 1 & 9.55 & 0.55 & 11.6 & 21.2 & \checkmark & \checkmark &  & \checkmark \\
high-dim/xgboost & 0.79 & 0.87 & 1 & 5.14 & 0.34 & 6.71 & 18.4 & \checkmark & \checkmark & \checkmark &  \\
jasmine/RF & 0.66 & 0.98 & 1 & 12.12 & 0.31 & 138 & 465 & \checkmark & \checkmark & \checkmark & \checkmark \\
jasmine/xgboost & 0.79 & 0.97 & 2 & 7.88 & 0.46 & 16.5 & 10.3 & \checkmark & \checkmark &  &  \\
madeline/RF & 0.87 & 0.88 & 1 & 3.12 & 0.17 & 4.03 & 13.9 & \checkmark &  & \checkmark &  \\
madeline/xgboost & 1.05 & 0.88 & 2 & 2.83 & 0.22 & 2.62 & 5.80 & \checkmark & \checkmark & \checkmark &  \\
volkert/RF & 0.26 & 0.91 & 2 & 7.07 & 0.82 & 21.1 & 10.9 & \checkmark & \checkmark &  &  \\
cifar10/wide\_resnet/2048 & 0.33 & 0.67 & 2 & 4.20 & 0.31 & 34.5 & 51.8 &  & \checkmark & \checkmark & \checkmark \\
cifar10/wide\_resnet/256 & 0.43 & 0.49 & 0 & 7.43 & 0.31 & 130 & 12.3 &  &  & \checkmark & \checkmark \\
cifar100/wide\_resnet/2048 & 0.41 & 0.68 & 1 & 2.69 & 0.37 & 22.3 & 10.7 &  &  & \checkmark &  \\
cifar100/wide\_resnet/256 & 0.38 & 0.61 & 0 & 3.92 & 0.74 & 18.2 & 15.6 &  &  &  & \checkmark \\
fashion\_mnist/max\_pooling\_cnn/2048 & 0.30 & 0.65 & 1 & 2.62 & 0.41 & 70.8 & 167 &  &  &  & \checkmark \\
fashion\_mnist/max\_pooling\_cnn/256 & 0.25 & 0.58 & 0 & 4.20 & 0.35 & 9.23 & 24.2 &  &  & \checkmark &  \\
fashion\_mnist/simple\_cnn/2048 & 1.01 & 0.61 & 2 & 3.48 & 0.33 & 29.3 & 116 & \checkmark & \checkmark & \checkmark & \checkmark \\
fashion\_mnist/simple\_cnn/256 & 1.12 & 0.54 & 0 & 4.75 & 0.20 & 126 & 34.9 & \checkmark &  & \checkmark & \checkmark \\
imagenet/resnet/256 & 0.28 & 0.67 & 0 & 3.41 & 0.46 & 11.0 & 7.37 &  &  &  &  \\
imagenet/resnet/512 & 0.56 & 0.58 & 0 & 2.90 & 0.44 & 27.2 & 9.68 &  &  &  & \checkmark \\
lm1b/transformer/2048 & 0.63 & 0.64 & 1 & 5.36 & 0.63 & 6.97 & 8.60 &  & \checkmark &  &  \\
mnist/max\_pooling\_cnn/2048 & 0.62 & 0.57 & 1 & 3.29 & 0.09 & 2.01 & 56.2 &  &  & \checkmark &  \\
mnist/max\_pooling\_cnn/256 & 0.71 & 0.39 & 0 & 4.67 & 0.13 & 48.5 & 241 &  &  & \checkmark & \checkmark \\
mnist/simple\_cnn/2048 & 0.56 & 0.56 & 0 & 3.08 & 0.42 & 672 & 54.3 &  &  &  & \checkmark \\
mnist/simple\_cnn/256 & 0.99 & 0.42 & 0 & 4.66 & 0.14 & 32.2 & 96.3 & \checkmark &  & \checkmark & \checkmark \\
svhn\_no\_extra/wide\_resnet/1024 & 0.20 & 0.49 & 1 & 4.89 & 0.68 & 4.97 & 21.4 &  & \checkmark &  &  \\
svhn\_no\_extra/wide\_resnet/256 & 0.15 & 0.40 & 0 & 7.03 & 0.70 & 71.4 & 51.5 &  &  &  & \checkmark \\
translate\_wmt/xformer\_translate/64 & 0.79 & 0.62 & 2 & 3.48 & 0.71 & 22.8 & 5.94 & \checkmark & \checkmark &  &  \\
uniref50/transformer/128 & 0.79 & 0.66 & 2 & 4.78 & 0.56 & 7.47 & 10.7 & \checkmark & \checkmark &  &  \\ \hline

\end{tabular}
\caption{Per-task flacco-style landscape features and the resulting membership in the four landscape axes used by Fig.~\ref{fig:landscape-radar}. Higher \texttt{disp\_05}, \texttt{ic\_hmax} indicate a more rugged landscape; higher \texttt{n\_pk}, \texttt{kurt} indicate multi-modality; lower \texttt{fdc} indicates a more deceptive landscape; larger \texttt{coef$_{\max/\min}$} and \texttt{quad.cond} indicate stronger anisotropy. The last four columns (\textbf{R}ugged, \textbf{M}ulti-Modal, \textbf{D}eceptive, \textbf{A}nisotropic) mark whether the task lies in the top 50\% of all 28 tasks by that combined landscape axis; membership is non-exclusive.}
\label{tab:landscape-features}
\end{table*}

\end{document}